\documentclass[10pt,twocolumn,letterpaper]{article}

\usepackage[pagenumbers]{iccv} % To force page numbers, e.g. for an arXiv version

\usepackage{pifont}

\usepackage{caption}
\newcommand{\framework}{\textbf{DreamRelation}\xspace}
\newcommand{\frameworkplain}{DreamRelation\xspace}
\newcommand{\tripletLoRAName}{relation LoRA triplet\xspace}
\newcommand{\trainStrategyName}{hybrid mask training strategy\xspace}
\newcommand{\trainStrategyNameShort}{HMT\xspace}
\newcommand{\CLName}{space-time relational contrastive loss\xspace}
\newcommand{\CLNameShort}{RCL\xspace}

\newlength\savewidth\newcommand\shline{\noalign{\global\savewidth\arrayrulewidth\global\arrayrulewidth1.25pt}\hline\noalign{\global\arrayrulewidth\savewidth}}

\newcommand{\tabincell}[2]{\begin{tabular}{@{}#1@{}}#2\end{tabular}}

\definecolor{darkred}{rgb}{0.7,0.1,0.1}
\definecolor{darkgreen}{rgb}{0.1,0.6,0.1}

\newcommand{\std}[1]{\footnotesize{#1}}
\usepackage{amsmath,amsfonts,bm}

\def\eqref#1{equation~\ref{#1}}
\def\1{\bm{1}}

\DeclareMathAlphabet{\mathsfit}{\encodingdefault}{\sfdefault}{m}{sl}
\SetMathAlphabet{\mathsfit}{bold}{\encodingdefault}{\sfdefault}{bx}{n}

\usepackage{wrapfig}

\definecolor{iccvblue}{rgb}{0.21,0.49,0.74}
\usepackage[pagebackref,breaklinks,colorlinks,allcolors=iccvblue]{hyperref}

\def\confName{ICCV}
\def\confYear{2025}

\title{
\framework: Relation-Centric Video Customization
}

\author{%
 Yujie Wei$^{1}$, Shiwei Zhang$^{2*}$, Hangjie Yuan$^2$, Biao Gong$^3$, Longxiang Tang$^2$, Xiang Wang$^2$, \\ Haonan Qiu$^4$, Hengjia Li$^5$, Shuai Tan$^3$, Yingya Zhang$^2$, Hongming Shan$^{1\dagger}$
 \\\\
 $^1$Fudan University\qquad$^2$Alibaba Group\qquad
 $^3$Ant Group
 \\
 $^4$Nanyang Technological University\qquad  $^5$Zhejiang University
 \\ 
 {\tt\small yjwei22@m.fudan.edu.cn},\quad
 {\tt\small zhangjin.zsw@alibaba-inc.com},\quad
 {\tt\small hmshan@fudan.edu.cn}\\
 \vspace{-0.6em} \\ 
 Project page: \url{https://dreamrelation.github.io}
}

\usepackage[misc]{ifsym}
\newcommand\blfootnote[1]{
    \begingroup
    \renewcommand\thefootnote{}\footnote{#1}
    \addtocounter{footnote}{-1}
    \endgroup
}
\usepackage[hang]{footmisc}

\begin{document}

\twocolumn[{
\renewcommand\twocolumn[1][]{#1}
\maketitle
\begin{center}
    \vspace{-14pt}
    \includegraphics[width=1.0\linewidth]{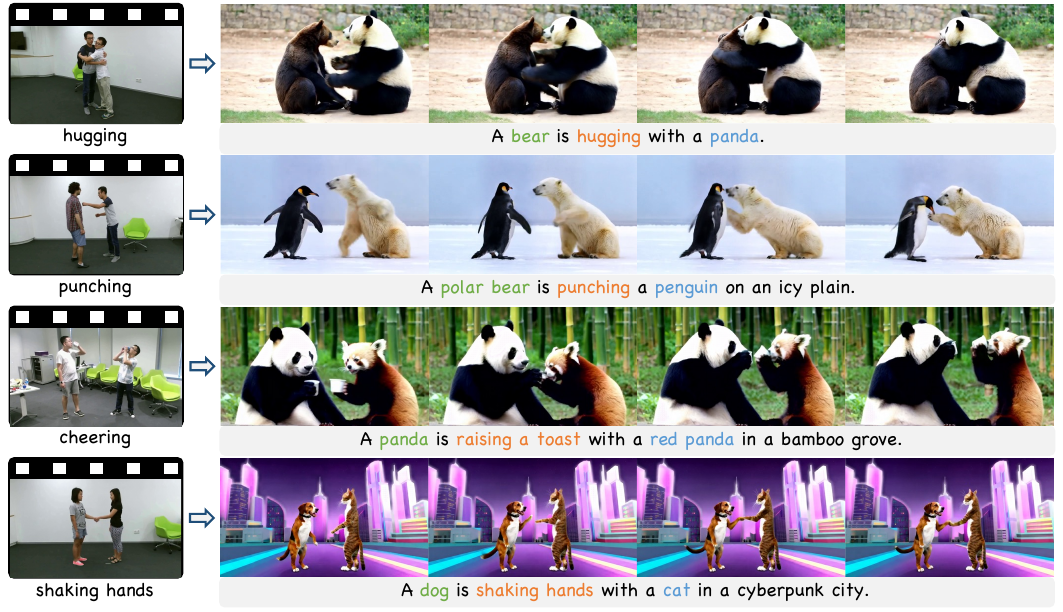}
    \vspace{-10pt}
    \captionsetup{type=figure}
    \caption{
        \textbf{Relational video customization results of \framework.} Given a few exemplar videos, our method can customize specific relations and generalize them to novel domains, where animals mimic human interactions.
    }
    \label{fig:teaser}
    \vspace{6pt}
\end{center}
}]
{
    \blfootnote{
        * Project Leader \\
        $^\dagger$ Corresponding Author 
        }
}

\begin{abstract}
Relational video customization refers to the creation of personalized videos that depict user-specified relations between two subjects, a crucial task for comprehending real-world visual content.
While existing methods can personalize subject appearances and motions, they still struggle with complex relational video customization, where precise relational modeling and high generalization across subject categories are essential.
The primary challenge arises from the intricate spatial arrangements, layout variations, and nuanced temporal dynamics inherent in relations; consequently, current models tend to overemphasize irrelevant visual details rather than capturing meaningful interactions.
To address these challenges, we propose \framework, a novel approach that personalizes relations through a small set of exemplar videos, leveraging two key components: Relational Decoupling Learning and Relational Dynamics Enhancement.
First, in Relational Decoupling Learning, we disentangle relations from subject appearances using \tripletLoRAName and \trainStrategyName, ensuring better generalization across diverse relationships.
Furthermore, we determine the optimal design of \tripletLoRAName by analyzing the distinct roles of the query, key, and value features within MM-DiT's attention mechanism, making \frameworkplain the first relational video generation framework with explainable components.
Second, in Relational Dynamics Enhancement, we introduce \CLName, which prioritizes relational dynamics while minimizing the reliance on detailed subject appearances.
Extensive experiments demonstrate that \frameworkplain outperforms state-of-the-art methods in relational video customization.
Code and models will be made publicly available.
\end{abstract}    
\section{Introduction}
\label{sec:intro}
Recent advancements in text-to-video (T2V) generation, particularly through powerful video diffusion transformers (DiT)~\cite{DiT, yang2024cogvideox, sora}, have significantly propelled customized video generation~\cite{wei2024dreamvideo, jeong2024vmc, yuan2024identity}.
While existing methods succeed in customizing subject appearances and single-object motions~\cite{wu2024customcrafter, motionDirector, wang2024motioninversion}, the challenging task of customizing higher-order interactions between subjects (\textit{i.e.}, Relational Video Customization) remains under-explored due to its intrinsic complexity.
Enhancing video generation through customized relations is crucial for real-world applications such as filmmaking, enabling a more profound comprehension and production of complex relational visual content.

\begin{figure}[t]
  \centering
   \includegraphics[width=1.0\linewidth]{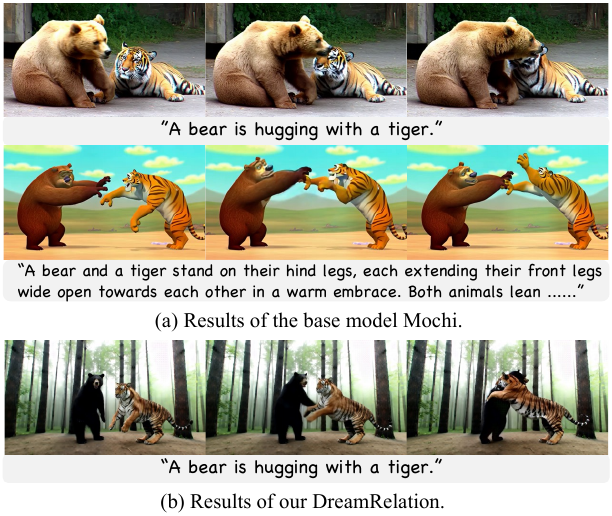}
   \caption{(a) General Video DiT models like Mochi~\cite{genmo2024mochi} often struggle to generate unconventional or counter-intuitive interactions, even with detailed descriptions.
   (b) Our method can customize a specific relation to generate videos on new subjects. 
   }
   \label{fig:base_model_relation}
   \vspace{-4pt}
\end{figure}

We formulate the task of Relational Video Customization as follows:
given exemplar videos representing a relational pattern $<${\color[HTML]{65a33e}subject}, {\color[HTML]{ea722b}relation}, {\color[HTML]{5190cf}subject}$>$, the model aims to generate videos that exhibit the specified relation within the pattern, as shown in Fig.~\ref{fig:teaser}.
While general text-to-video DiTs like Mochi~\cite{genmo2024mochi} can generate videos depicting certain relational concepts, they often fail to:
(1) produce unconventional or counter-intuitive interactions, such as animals engaging in human-like relationships as illustrated in Figs.~\ref{fig:base_model_relation}, even when provided with detailed prompts;
(2) generate videos that adhere to precise relational dynamics, such as \textit{``two people approaching each other from predefined positions.''}
These issues highlight the need for a novel video generation method to precisely customize desired relations.

\begin{figure}[t]
  \centering
   \includegraphics[width=1.0\linewidth]{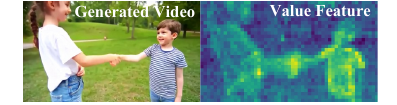}
   \caption{Averaged value feature across all layers and frames in Mochi. We identify that the relations encompass intricate spatial arrangements, layout variations, and nuanced temporal dynamics, presenting challenges in relational video customization.}
   \label{fig:base_model_value_feature}
   \vspace{-4pt}
\end{figure}

A straightforward approach involves adapting existing video subject or motion customization methods to customize relations between subjects. 
However, while subject customization techniques like Dreamix~\cite{dreamix} capture detailed appearances using low-level reconstruction loss, they may hinder high-level relation learning due to severe appearance leakage.
Similarly, motion customization methods such as MotionInversion~\cite{wang2024motioninversion} excel in transferring single-object motions but struggle to precisely capture relational dynamics between two subjects.
We identify that the key challenge stems from the complexity inherent in the relations, which involve intricate 
\textit{spatial arrangements}, \textit{layout variations}, and \textit{nuanced temporal dynamics}.
To illustrate this, we visualize the Value features in Fig.~\ref{fig:base_model_value_feature} and provide detailed analysis in Sec.~\ref{sec:model_analysis}.
This tangled nature may prevent accurate modeling of relations and cause models to focus on irrelevant subject appearances.
This raises a critical research question:
\textit{How can we {decouple relations and subject appearances} while accurately {modeling relational dynamics} to enhance generalizability?}

To that end, we propose \emph{\framework}, a relational video customization method that personalizes user-specified relations from exemplar videos through two concurrent processes: relational decoupling learning and relational dynamics enhancement.
In relational decoupling learning, we decompose the relational pattern from input videos into relational and appearance information using devised \tripletLoRAName, a composite LoRA~\cite{hu2021lora} set comprising relation LoRA sets and subject LoRA sets.
To facilitate this decoupling, we introduce \trainStrategyName that guides the two types of LoRAs to focus on designated regions with corresponding masks, achieved by a LoRA selection strategy and an enhanced diffusion loss based on masks to amplify the learning in target areas.

Furthermore, building on the MM-DiT~\cite{SD3} architecture, we analyze the query, key, and value features within the full attention, and empirically identify that the query, key, and value matrices serve distinct roles in the relation customization task. 
This insight motivates our design of \tripletLoRAName, particularly in determining the optimal placement of LoRA components within the model architecture to maximize relational customization effectiveness.

To explicitly enhance relational dynamics learning,
we propose a novel \CLName, which emphasizes relational dynamics while reducing the focus on detailed appearances during training. 
Concretely, we pull relational dynamics representations closer through frame differences in model outputs of videos depicting the same relation, while distancing them from appearance representations derived from single-frame outputs.

We curate a dataset comprising 26 human interactions from publicly available action recognition datasets~\cite{shahroudy2016ntu, liu2019ntu} to comprehensively evaluate relational video customization. 
Each video is annotated with a textual prompt, and approximately 20 videos per relation type are randomly selected for training.
The evaluation is conducted on diverse subjects using 40 designed textual prompts.
Extensive experimental results demonstrate that our \frameworkplain outperforms state-of-the-art methods in this task.

Our contributions are summarized as follows:
\begin{itemize}
\item 
We make the first attempt at the Relational Video Customization task by presenting \framework, a method that generates videos depicting customized relations based on the MM-DiT architecture.
\item 
We devise \tripletLoRAName with \trainStrategyName to explicitly decouple relation and subject appearances. 
To determine the optimal model design of our method, we further analyze the roles of query, key, and value features in MM-DiT full attention.
\item We propose a novel \CLName to enhance relation learning by emphasizing relational dynamics while reducing focus on appearances.
\item Extensive experimental results demonstrate that \frameworkplain achieves state-of-the-art performance on relational video customization.
\end{itemize}

\section{Related Work}
\label{sec:related_work}

\noindent\textbf{Text-to-video diffusion models.}\quad
Text-to-video generative models have achieved breakthroughs in generating high-quality and diverse videos using textual prompts~\cite{animatediff, esser2023structure, imagenVideo, svd, bar2024lumiere, kondratyuk2023videopoet, wang2023lavie, show1, latent_shift, CDT, yuan2024instructvideo, qing2024hierarchical, liu2024timestep, wang2023videocomposer, wang2023videolcm, wang2024recipe, wang2024unianimate, tan2024animate, tan2024mimir, tan2025edtalk, tan2024flowvqtalker}.
VDM~\cite{VDM} introduces diffusion models into video generation by modeling video distribution in pixel space.
ModelScopeT2V~\cite{modelScope} and VideoCrafter~\cite{chen2023videocrafter1, chen2024videocrafter2} integrate spatiotemporal blocks for text-to-video generation.
With the success of DiT~\cite{DiT} that introduces Transformers~\cite{vaswani2017attention} as the backbone of diffusion models, the generated video quality has improved with increased parameters~\cite{sora, Open-Sora, Open-Sora-Plan, ma2024latte, Vchitect}.
CogVideoX~\cite{yang2024cogvideox} incorporates 3D VAE and expert transformers, enhancing video coherence.
Mochi~\cite{genmo2024mochi} proposes an Asymmetric Diffusion Transformer architecture to scale parameters.
HunyuanVideo~\cite{kong2024hunyuanvideo} enhances architecture design and model training, achieving leading performance.
These advancements pave the way for relational video customization.

\noindent\textbf{Customized video generation.}\quad
Building upon achievements in image generation and personalization~\cite{DDPM, stableDiffusion, podell2023sdxl, zhang2023adding, textInversion, dreambooth, wei2023elite, SuTI, dalva2024noiseclr, xu2024facechain, zhou2024storydiffusion, cao2023masactrl}, customized video generation has garnered growing attention~\cite{dreamix, chefer2024still_moving, ma2024magicme, he2024id_animator}.
Many studies focus on generating personalized videos using a few subject or facial images~\cite{yuan2024identity, wei2024dreamvideo, wu2024motionbooth, zhou2024sugar, wu2024videomaker, zhang2025fantasyid, she2025customvideox, wu2024customcrafter, wei2024dreamvideo2, li2024personalvideo}, while others tackle the challenging multi-subject video customization~\cite{chen2023videodreamer, wang2024customvideo, chen2024disenstudio, chen2025multi, huang2025conceptmaster}.
Besides subject customization, motion customization or motion transfer have also gained significant interest~\cite{motionDirector, jeong2024vmc, customize_a_video, yatim2024space, tu2024motionfollower, tu2024motioneditor, xu2024combo, jeong2024dreammotion}. 
For example, MotionInversion~\cite{wang2024motioninversion} integrates motion embeddings into the temporal attention of video diffusion models to learn motion dynamics.
While these methods effectively capture the subject appearances or single-object motions, the challenging task of customizing interactions between two subjects remains underexplored due to its inherent complexity.
In this work, we pioneer this relational video customization task by presenting \frameworkplain, which can personalize specific relations and generate diverse videos aligned with text prompts.

\noindent\textbf{Relation generation.}\quad
Early works on relational image generation focus on human-object interactions using additional conditions like bounding boxes~\cite{hua2021exploiting, gao2020interactgan, hoe2024interactdiffusion}.
Recently, inspired by image customization methods, several works have explored relational image customization to personalize user-specific interactions from a few relational images~\cite{huang2024reversion, ge2024customizing, shi2024relationbooth}.
For instance, ReVersion~\cite{huang2024reversion} utilizes inversion techniques to capture relational information in the text embedding space. 
Despite these advancements, existing methods are confined to the relatively simple relations depicted in images.
Direct adaptation of these image-based methods for relational video customization often leads to inaccurate relation modeling since dynamic and sequential interactions cannot be fully represented in a single image.
In contrast, we design our method based on Video DiT architecture and precisely model relations through relational decoupling learning and relational dynamics enhancement.

\begin{figure*}[t]
  \centering
   \includegraphics[width=1.0\linewidth]{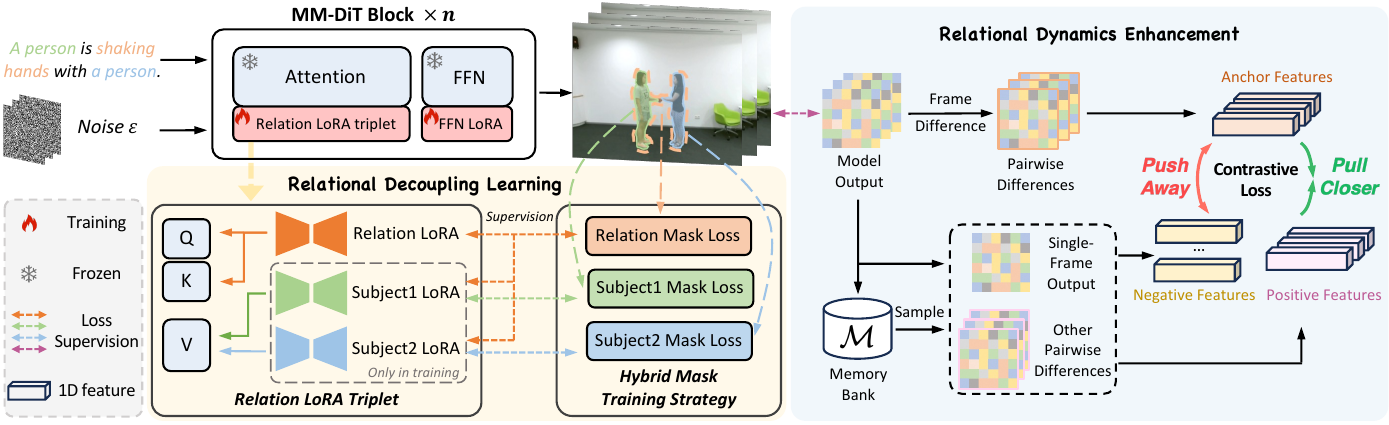}
   \caption{\textbf{Overall framework of \framework}.
   Our method decomposes relational video customization into two concurrent processes.
   (1) In Relational Decoupling Learning, Relation LoRAs in \tripletLoRAName capture relational information, while Subject LoRAs focus on subject appearances. This decoupling process is guided by \trainStrategyName based on their corresponding masks.
   (2) In Relational Dynamics Enhancement, the proposed \CLName pulls relational dynamics features (anchor and positive features) from pairwise differences closer, while pushing them away from appearance features (negative features) of single-frame outputs.
   During inference, subject LoRAs are excluded to prevent introducing undesired appearances and enhance generalization.
   }
   \label{fig:framework}
\end{figure*}
\section{\frameworkplain}
\label{sec:method}
Our \frameworkplain aims to generate videos depicting a specified relation expressed in a few exemplar videos while aligning with textual prompts, as illustrated in Fig.~\ref{fig:framework}.
We begin by introducing preliminaries in Sec.~\ref{sec:preliminaries}.
We then detail relational decoupling learning and relational dynamics enhancement in Secs.~\ref{sec:relation_decouple} and~\ref{sec:CL}, respectively, along with an analysis of the query, key, and value features in Sec.~\ref{sec:model_analysis}.

\subsection{Preliminaries of Video DiT}
\label{sec:preliminaries}
Text-to-video diffusion transformers~(DiTs) show growing attention due to their capacity to generate high-fidelity, diverse, and long-duration video. Current Video DiTs~\cite{yang2024cogvideox, genmo2024mochi} predominantly adopt MM-DiT~\cite{SD3} architecture with full attention and employ diffusion processes~\cite{DDPM} in latent space with a 3D VAE~\cite{vae}.
Given latent code $\bm{z}_0 \in \mathbb{R}^{f \times h \times w \times c}$ from video data $\bm{x}_0 \in \mathbb{R}^{F \times H \times W \times 3}$ with its textual prompt $\bm{c}$, the optimization process is defined as:
\begin{equation}
    \mathcal{L}(\theta) = \mathbb{E}_{\bm{z}, \epsilon, \bm{c}, t} \big[\left\| \epsilon - \epsilon_{\theta}(\bm{z}_t, \bm{c}, t) \right\|_{2}^{2}\big],
\end{equation}
where $\epsilon \in \mathcal{N}(0,1)$ is random noise from a Gaussian distribution, and $\bm{z}_t$ is a noisy latent code at timestep $t$ based on $\bm{z}_0$ with the predefined noise schedule.
In this work, we choose Mochi~\cite{genmo2024mochi} as our base Video DiT model.

\subsection{Relational Decoupling Learning}
\label{sec:relation_decouple}
\noindent\textbf{Relation LoRA triplet.}\quad
\label{sec:triple_lora}
To customize complex relations between subjects, we decompose the relational pattern from exemplar videos into distinct components emphasizing subject appearances and relations.
Formally, given a few videos depicting interactions between two subjects, we represent their relational patterns as a triplet $<${\color[HTML]{65a33e}subject}, {\color[HTML]{ea722b}relation}, {\color[HTML]{5190cf}subject}$>$, denoted as $<${\color[HTML]{65a33e}$S_1$}, {\color[HTML]{ea722b}$R$}, {\color[HTML]{5190cf}$S_2$}$>$ for brevity, where $S_1$ and $S_2$ are two subjects and $R$ is the relation~\cite{yuan2022rlip}.

To differentiate relations and subject appearances in the relational pattern,
we introduce \tripletLoRAName, a composite LoRA set comprising Relation LoRAs to model relational information and two Subject LoRAs to capture appearance information, as depicted in Fig.~\ref{fig:framework}.
Specifically, we inject Relation LoRAs into the query and key matrices of the MM-DiT full attention.
Concurrently, we design two Subject LoRAs corresponding to the two subjects involved in the relation and inject them into the value matrix. 
This design is motivated by our empirical findings that the query, key, and value matrices serve distinct roles within the MM-DiT full attention. 
More details on the analysis are provided in Sec.~\ref{sec:model_analysis}. 
Additionally, we devise an FFN LoRA to refine the outputs of the Relation and Subject LoRAs and inject it into the linear layers of full attention.
Note that the two branches of text and vision tokens in MM-DiT are processed by different LoRA sets.

\begin{figure*}[t]
  \centering
   \includegraphics[width=1.0\linewidth]{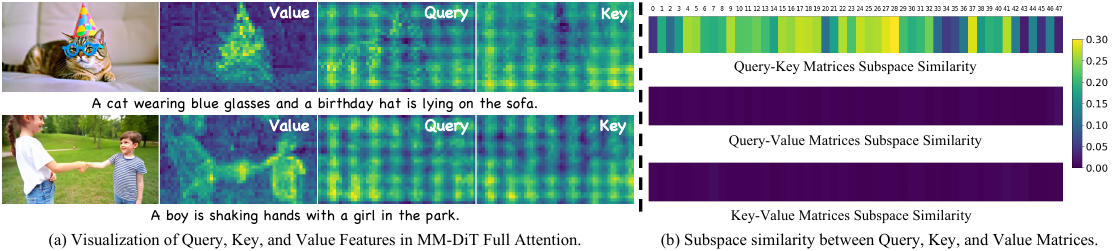}
   \caption{\textbf{Features and subspace similarity analysis of MM-DiT}. (a) Value features across different videos encapsulate rich appearance information, and relational information often intertwines with these appearance cues. Meanwhile, query and key features exhibit similar patterns that differ from those of value features.
   (b) We perform singular value decomposition on the query, key, and value matrices of each MM-DiT block and compute the similarity of the subspaces spanned by their top-k left singular vectors, indicating query and key matrices share more common information while remaining independent of the value matrix.}
   \label{fig:model_analysis}
\end{figure*}

\noindent\textbf{Hybrid mask training strategy.}\quad
\label{sec:train_strategy}
To achieve the decoupling of relational and appearance information in the introduced \tripletLoRAName,
we propose \trainStrategyName (\trainStrategyNameShort) to guide Relation and Subject LoRAs to focus on designated regions using corresponding masks.
We first employ Grounding DINO~\cite{groundingdino} and SAM~\cite{sam} to derive masks for the two individuals in a video, indicated as Subject Masks $M_{S_1}$ and $M_{S_2}$.
Inspired by representative relation detection approaches~\cite{tang2020unbiased-SGG,xu2017SGG-message-passing,yang2018graph-rcnn} that utilize minimum enclosing rectangles to delineate subject-object interaction zones, we define the Relation Mask $M_R$ as the union of the two Subject Masks to indicate the relation area.
Since the 3D VAE in Video DiT compresses the video's temporal dimensions by a factor of $T_c$, we average the masks over every $T_c$ frame to represent the latent masks.

We then devise a LoRA selection strategy and an enhanced diffusion loss for better disentanglement during training.
Specifically, we randomly select either the Relation LoRAs or one type of Subject LoRAs in \tripletLoRAName to update for each training iteration. 
When the Relation LoRAs are chosen, the two Subject LoRAs are trained simultaneously to provide appearance cues, assisting the Relation LoRAs in concentrating on relational information.
This process facilitates the decoupling of relational and appearance information.
The FFN LoRAs are consistently engaged throughout training to refine outputs from the selected Relation or Subject LoRAs. 

Following LoRA selection, we apply the corresponding masks to amplify the loss weight within the focused area, which can be defined as:
\begin{equation}
    \mathcal{L}_\text{rec} = \mathbb{E}_{\bm{z}, \epsilon, \bm{c}, t} 
    \big( \lambda_{m}\mathbf{M}_l
    + 1 \big)
    \cdot
    \big\| 
    \epsilon - \epsilon_{\theta}(\bm{z}_t, \bm{c}, t)
    \big\|_{2}^{2},
\label{eq:box_mask_loss}
\end{equation}
where $l \in \{S_1, S_2, R\}$ indicates the selected mask type, and $\lambda_{m}$ is the mask weight. 
By employing the LoRA selection strategy and the enhanced diffusion loss, Relation and Subject LoRAs are encouraged to concentrate on their designated area, facilitating effective relation customization and improving the generalization capacity.

\noindent\textbf{Inference.}\quad
During inference, we exclude Subject LoRAs to prevent undesired appearances and inject only Relation LoRAs and FFN LoRAs into the base Video DiT to maintain learned relations and enhance generalization.

\subsection{Analysis on Query, Key, and Value Features}
\label{sec:model_analysis}
To determine the optimal model design of our method,
we analyze the roles of query, key, and value features or matrices in MM-DiT's full attention through visualization and singular value decomposition, revealing their impacts on relational video customization.

\noindent\textbf{Visualization analysis.}\quad
We start with two types of videos: a single-subject video with multiple attributes, and a two-subject interaction video, as illustrated in Fig.~\ref{fig:model_analysis}(a). 
We compute the averaged query, key, and value features across all layers and attention heads at timestep 60, focusing solely on those associated with vision tokens.
These features are then reshaped into an $f\times h \times w$ format, and we visualize the averaged features across all frames with shape $h \times w$.
From the observations in Fig.~\ref{fig:model_analysis}(a), we draw two conclusions:

1) \textit{Value features across different videos encapsulate rich appearance information, and relational information often intertwines with these appearance cues}.
For instance, in the single-subject video, high-value feature responses occur at locations like ``blue glasses'' and ``birthday hat.'' 
In the two-subject video, high values are observed both in regions of relations (\eg, handshakes) and appearances (\eg, human face and clothing), indicating the entanglement of relational and appearance information within the features.

2) \textit{Query and key features exhibit highly abstract yet similar patterns, distinctly diverging from the value features}.
Unlike the obvious appearance information in value features, query, and key features exhibit homogeneity across different videos, clearly differing from value features. 
To further validate this point, we analyze query, key, and value matrices from a quantitative perspective.

\noindent\textbf{Subspace similarity analysis.}\quad
We further analyze the similarity of the subspace spanned by the singular vectors of the query, key, and value matrix weights from the base Video DiT model Mochi.
This similarity reflects the degree of overlap in contained information between two matrices.
For the query and key matrices, we apply singular value decomposition to obtain left-singular unitary matrices $U_Q$ and $U_K$.
Following~\cite{hu2021lora, liu2024towards}, we select the top $r$ singular vectors from $U_Q$ and $U_K$, and measure their normalized subspace similarity based on the Grassmann distance~\cite{hamm2008grassmann} using
$\frac{1}{r}\left\|U_Q^{r \top} U_K^{r}\right\|_F^2$.
The other similarities are calculated in a similar way.
The results in Fig.~\ref{fig:model_analysis}(b) demonstrate that the subspaces of the query and key matrices are highly similar, whereas their similarity to the value matrix is minimal.
This suggests that \textit{the query and key matrices in MM-DiT share more common information while remaining largely independent of the value matrix}.
In other words, the query and key matrices exhibit a strongly non-overlapping relationship with the value matrix, which facilitates the design of our decoupling learning.
This finding is consistent with the visualization results in Fig.~\ref{fig:model_analysis}(a).

Building on these observations, we empirically argue that the query, key, and value matrices serve distinct roles in relational video customization,  motivating our design of \tripletLoRAName.
Specifically, given that value features are rich in appearance information, we inject Subject LoRAs into the value matrix to focus on learning appearances.
In contrast, due to the homogeneity observed in the query and key features and their non-overlapping nature with the value matrix, which facilitates decoupling learning, we inject Relation LoRAs into both the query and key matrices to better disentangle relations from appearances.
The experimental results in Tab.~\ref{tab:ablation_relationLoRA_pos} confirm our analysis and verify that this design achieves the best performance.
We believe our findings can advance research in video customization based on MM-DiT architecture.

\subsection{Relational Dynamics Enhancement}
\label{sec:CL}
To explicitly enhance relational dynamics learning, we propose a novel \CLName (\CLNameShort), which emphasizes relational dynamics while reducing the focus on detailed appearance during training.
Specifically, at each timestep $t$, we compute the pairwise differences of the model output along the frame dimension, denoted as $\bar{\epsilon} \in \mathbb{R}^{(f-1) \times h \times w \times c}$.
We then reduce dependency on pixel-level information by averaging these differences across the spatial dimensions, resulting in 1D relational dynamics features $A \in \mathbb{R}^{(f-1) \times c}$, which serve as anchor features.
Subsequently, we sample $n_\text{pos}$ 1D relational dynamics features from other relation videos as positive samples $P \in \mathbb{R}^{(f-1) \times n_\text{pos} \times c}$.
For each frame in $A$, we sample $n_\text{neg}$ 1D features from single-frame model outputs $\epsilon_i \in \mathbb{R}^{1 \times h \times w \times c}$ as negative samples $N \in \mathbb{R}^{(f-1) \times n_\text{neg} \times c}$, which capture appearance information while excluding relational dynamics.

Our objective is to learn representations with relational dynamics by pulling together the pairwise differences from different videos depicting the same relation, while distancing them from spatial features of single-frame outputs to mitigate appearance and background leakage.
Following InfoNCE~\cite{miech2020end, oord2018representation} loss, we formulate the proposed loss as:
\begin{equation}
\mathcal{L}_{\text {RCL}}=\log \sum\limits_{i=1}^{f-1} 
\frac{- \sum\limits_{j=1}^{n_\text{pos}} \text{exp}(\frac{{A_i^{\top} P_{ij}}}{\tau})}
{\sum\limits_{j=1}^{n_\text{pos}} 
\text{exp}(\frac{{A_i^{\top} P_{ij}}}{\tau}) +\sum\limits_{k=1}^{n_\text{neg}} 
\text{exp}(\frac{{A_i^{\top} N_{ik}}}{\tau})},
\end{equation}
where $\tau$ is the temperature hyper-parameter.

Additionally, we maintain a memory bank $\mathcal{M}$ to store and update the positive and negative samples.
Both positive and negative samples are randomly selected from the 1D features of current batch videos and previously seen videos. This online dynamic update strategy can enlarge the number of positive and negative samples, enhancing the contrastive learning effect and training stability.
At each iteration, we store all current anchor features $A$ and the 1D features of $\epsilon_i$ into $\mathcal{M}$. The memory bank is implemented as a First In, First Out (FIFO) queue.

Overall, the training loss $\mathcal{L}_\text{total}$ consists of both reconstruction and contrastive learning loss, defined as:
\begin{equation}
    \mathcal{L}_\text{total} = \mathcal{L}_\text{rec} + \lambda_1 \mathcal{L}_\text{RCL},
\end{equation}
where $\lambda_1$ is the loss balancing weight.

\section{Experiment}
\label{sec:exp}
\begin{figure*}[t]
  \centering
   \includegraphics[width=1.0\linewidth]{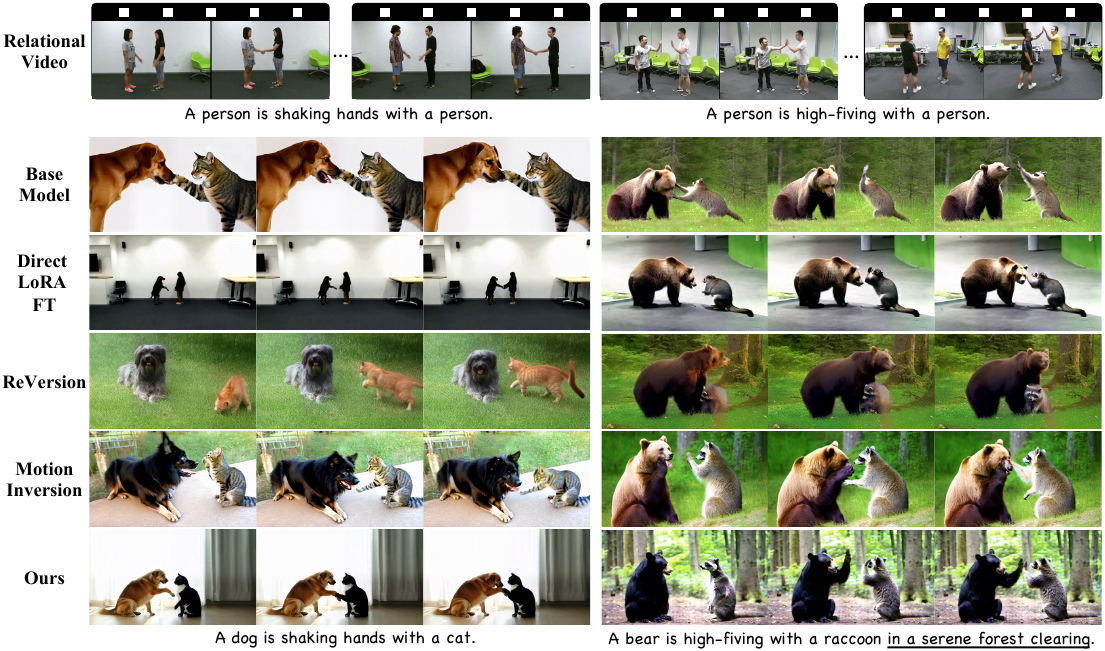}
   \caption{\textbf{Qualitative comparison results}. Our method outperforms all baselines in precisely capturing the intended relation and mitigating appearance and background leakage.}
   \label{fig:qualitative_compare_baselines}
\end{figure*}
\subsection{Experimental Setup}
\label{sec:exp_setup}
\noindent\textbf{Datasets.}\quad
We conduct experiments on the NTU RGB+D Action Recognition Dataset~\cite{shahroudy2016ntu, liu2019ntu}. 
We select 26 types of human relations, such as handshakes and hugs,
each labeled with a text prompt like ``A person is shaking hands with a person.''
For evaluation, we design 10$\times$26 prompts with uncommon subject interactions, such as ``A dog is shaking hands with a cat'', to assess generalization to novel domains.
More details are provided in
Appendix~\ref{app:exp_setup}.

\noindent\textbf{Baselines.}\quad
Given the absence of existing methods for relational video customization, we define four baseline categories: 
\textit{1)} The base model Mochi.
\textit{2)} Direct LoRA finetuning.
\textit{3)} Adapted relational image customization methods. We reproduce ReVersion~\cite{huang2024reversion} on Mochi for relational video customization. 
\textit{4)} Motion customization methods,
which mostly rely on Temporal Attention Layers that are absent in MM-DiT, face challenges in direct adaptation.
Thus, we choose the recent and adaptable MotionInversion~\cite{wang2024motioninversion} as a baseline, reproducing it on Mochi for comparison.

\noindent\textbf{Evaluation metrics.}\quad
We evaluate our method by focusing on four aspects:
\textit{1)} Relation Accuracy. 
Instead of using biased classifiers trained on test sets with limited diversity like previous methods~\cite{huang2024reversion, ge2024customizing}, which hinders test accuracy and generalizability, 
we propose the Relation Accuracy metric 
to assess relations using \textit{advanced Vision-Language Models (VLMs)}.
Specifically, we input generated videos to Qwen-VL-Max~\cite{Qwen-VL}, a leading VQA model, asking if the video matches the specified relation, and convert the yes/no responses into a relation accuracy percentage. 
We repeat this process 10 times to calculate the average accuracy.
\textit{2)} Text Alignment. 
We employ CLIP image-text similarity (CLIP-T) to measure 
alignment with text prompts.
\textit{3)} Temporal Consistency, which computes the average cosine similarity across consecutive frames~\cite{esser2023structure}.
\textit{4)} Video Quality. We use FVD to evaluate the video quality.
The reference videos are 800 videos from the AnimalKingdom test dataset~\citep{Animal_Kingdom}.

\noindent\textbf{Implementation details.}\quad
We adopt Mochi~\cite{genmo2024mochi} as our base model.
During training, we use AdamW~\citep{adamw} optimizer with a learning rate of 2e-4. The weight decay is set to 0.01, and the training iteration is 2400. 
We set LoRA rank to 16, $\lambda_{m}$ to 50, and $\lambda_1$ to 0.01. 
The resolution of generated videos is 61$\times$480$\times$848, and
the batch size is 1.
We set $n_{\text{pos}}$ to 4 and $n_{\text{neg}}$ to 10. The memory bank size is set to 64, and $\tau$ is 0.07.
During inference, we generate 30-fps videos using Mochi's default Euler Discrete method~\citep{liu2022flow, lipman2022flow} with 64 steps.
The classifier-free guidance~\citep{ho2022classifier_free_guide} scale is 6.0.

\begin{table}[t]
    \centering
    \vspace{-4mm}
    \resizebox{\columnwidth}{!}{
    \begin{tabular}{lcccc} 
        \textbf{Method} & \tabincell{c}{\textbf{Relation}\\\textbf{Accuracy}} & 
        \textbf{CLIP-T} & \tabincell{c}{\textbf{Temporal}\\\textbf{Consistency}} & 
        \textbf{FVD$\downarrow$} \\ 
        \shline
        \noalign{\smallskip}
         Mochi (base model)~\cite{genmo2024mochi} & 0.2623\std{$\pm$0.04} & 0.3237 & 0.9888 & \textbf{2047.37} \\
         Direct LoRA finetuning & 0.3258\std{$\pm$0.05} & 0.2966 & 0.9945 & 2229.08 \\
         ReVersion~\cite{huang2024reversion} & 0.2690\std{$\pm$0.01} & 0.3013 & 0.9921 & 2682.69 \\
         MotionInversion~\cite{wang2024motioninversion} & 0.3151\std{$\pm$0.03} & 0.3217 & 0.9855 & 2084.51 \\
         \hline
         \noalign{\smallskip}
         \textbf{\framework} & \textbf{0.4452}\std{$\pm$0.01} & \textbf{0.3248} & \textbf{0.9954} & 2079.87
         \\ 
    \end{tabular}
    }
    \caption{Quantitative comparison results.
    }
    \vspace{-6mm}
    \label{tab:compare}
\end{table}

\subsection{Main Results}

\noindent\textbf{Qualitative results.}\quad
Qualitative comparisons in Fig.~\ref{fig:qualitative_compare_baselines}
reveal that all baseline methods, including the base model Mochi, fail to generate videos that match the relations defined in exemplar videos. 
For example, Direct LoRA finetuning struggles with appearance and background leakage, while other methods like MotionInversion cannot capture desired relational dynamics due to the complexity inherent in relations.
In contrast, our \frameworkplain precisely generates videos with intended relations and diverse subjects, effectively preventing appearance and background leakage.

\noindent\textbf{Quantitative results.}\quad
Tab.~\ref{tab:compare} presents the quantitative comparison results.
Direct LoRA finetuning improves the base model's Relation Accuracy but suffers from reduced CLIP-T and FVD due to appearance leakage.
Inversion-based methods like ReVersion and MotionInversion achieve better CLIP-T than finetuning but fail to model desired relations accurately.
In contrast, while comparable to the base model in FVD, our \frameworkplain consistently surpasses baselines across other metrics, 
verifying its effectiveness.

\noindent\textbf{Attention map analysis.}\quad
\begin{figure}[t]
  \centering
  \vspace{-4mm}
   \includegraphics[width=1.0\linewidth]{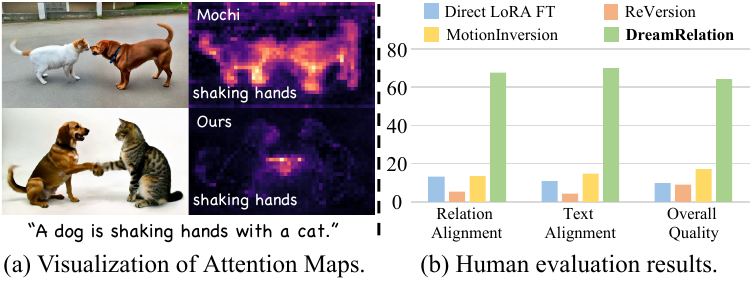}
   \caption{(a) Our method focuses on the desired relational region. (b) Our method is most preferred by users across all aspects.}
   \vspace{-6mm}
   \label{fig:exp_attn_map_and_user_study}
\end{figure}
To verify the effectiveness of our method, we compute averaged attention maps from all layers and heads, extracting values for text tokens of relations like ``shaking hands'' and all vision tokens
~\cite{cai2024ditctrl}. 
These attention maps are reshaped and visualized in Fig.~\ref{fig:exp_attn_map_and_user_study}(a).
We observe that 
the base model's attention map for ``shaking hands'' is messy, leading to poor generation.
In contrast, our method's attention map effectively focuses on the relational area, producing more natural results and demonstrating its capability to capture relational information.

\noindent\textbf{User study.}\quad
We conduct user studies to evaluate our \frameworkplain, involving 15 annotators who rate 180 video groups generated by four methods.
Each group contains four generated videos, a reference video, and a textual prompt.
Evaluations are based on majority votes in three aspects:
Relation Alignment, Text Alignment, and Overall Quality.
Results in Fig.~\ref{fig:exp_attn_map_and_user_study}(b) indicate that our method is most preferred by users across all aspects.
More details about the user study are provided in Appendix~\ref{app:user_study}.

\begin{figure}[t]
  \centering
   \includegraphics[width=1.0\linewidth]{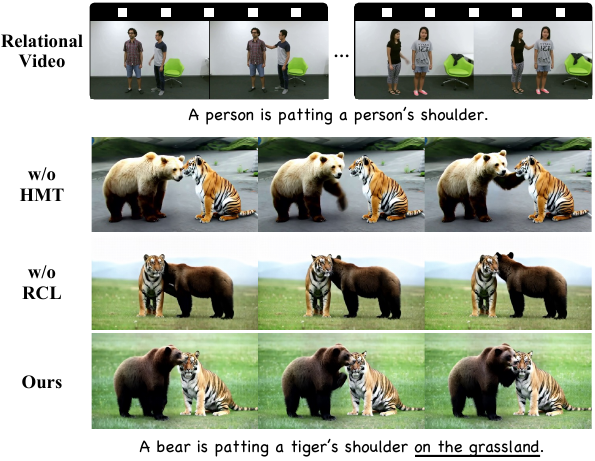}
   \caption{\textbf{Qualitative ablation study on each component.}}
   \label{fig:qualitative_ablation}
\end{figure}
\begin{table}[t]
    \centering
    \vspace{-2mm}
    \resizebox{\columnwidth}{!}{
    \begin{tabular}{lcccc} 
        \textbf{Method} & \tabincell{c}{\textbf{Relation}\\\textbf{Accuracy}} & \textbf{CLIP-T} & \tabincell{c}{\textbf{Temporal}\\\textbf{Consistency}} & 
        \textbf{FVD$\downarrow$} \\ 
        \shline
        \noalign{\smallskip}
         w/o \trainStrategyNameShort & 0.3574\std{$\pm$0.02} & 0.3244 & 0.9938 & 2248.52 \\
         w/o \CLNameShort & 0.3416\std{$\pm$0.03} & 0.3185 & 0.9953 & 2136.95 \\
         \hline
         w/o Relation LoRAs  &  0.3626\std{$\pm$0.02} & 0.3035 & 0.9950 & 2318.49 \\
         w/o Subject LoRAs & 0.3769\std{$\pm$0.04} & 0.3147 & 0.9949 & 2408.59 \\
         w/o FFN LoRAs &  0.4021\std{$\pm$0.03} & 0.3241 & 0.9914 & 2369.98 \\
         \hline
         \noalign{\smallskip}
         \textbf{ours} & \textbf{0.4452}\std{$\pm$0.01} & \textbf{0.3248} & \textbf{0.9954} & \textbf{2079.87} \\ 
    \end{tabular}
    }
    \caption{Ablation studies on effects of \trainStrategyName (\trainStrategyNameShort), \CLName (\CLNameShort), and each type of LoRA. Removing any of the above components significantly reduces the overall performance.
    }
    \vspace{-4mm}
    \label{tab:ablation_component_each_LoRA_type}
\end{table}

\subsection{Ablation Studies}

\noindent\textbf{Ablation on each component.}\quad
We perform an ablation study on the effects of each component, as shown in Fig.~\ref{fig:qualitative_ablation}. 
Without \trainStrategyName, the model generates the desired relations but experiences some background leakage due to incomplete decoupling of relational and appearance information.
Omitting \CLName reduces background leakage but results in videos exhibiting inaccurate relations.

Quantitative results in Tab.~\ref{tab:ablation_component_each_LoRA_type} show that removing \trainStrategyName or \CLName degrades performance across all metrics,
confirming that each component is crucial to overall performance;
see Appendix~\ref{app:more_ablation} for more ablation studies.

\noindent\textbf{Ablation on  each LoRA in \tripletLoRAName.}\quad
We conduct ablation studies to verify each LoRA's effects.
The results in Tab.~\ref{tab:ablation_component_each_LoRA_type} indicate that 
removing Relation LoRAs or Subject LoRAs significantly reduces Relation Accuracy and CLIP-T due to insufficient decoupling of appearance and relational information. 
Excluding FFN LoRAs also lowers accuracy, highlighting the need for refinement.

\noindent\textbf{Ablation on  Relation LoRAs position.}\quad
\begin{table}[t]
    \centering
    \resizebox{\columnwidth}{!}{
    \begin{tabular}{cc|cccc} 
        \tabincell{c}{\textbf{Relation}\\\textbf{LoRA}} & \tabincell{c}{\textbf{Subject}\\\textbf{LoRA}} & \tabincell{c}{\textbf{Relation}\\\textbf{Accuracy}} &
        \textbf{CLIP-T} & \tabincell{c}{\textbf{Temporal}\\\textbf{Consistency}} & 
        \textbf{FVD$\downarrow$} \\ 
        \shline
        \noalign{\smallskip}
        V & Q, K & 0.3444\std{$\pm$0.02} & 0.3225 & 0.9953 & 2233.48 \\
        Q & K, V & 0.3921\std{$\pm$0.03} & \textbf{0.3301} & 0.9951 & 2284.65 \\
        K, V & Q & 0.3937\std{$\pm$0.04} & 0.3196 & \textbf{0.9954} & 2180.27 \\
        \hline
        \noalign{\smallskip}
        Q, K & V & \textbf{0.4452}\std{$\pm$0.01} & 0.3248 & \textbf{0.9954} & \textbf{2079.87} \\
    \end{tabular}
    }
    \caption{Ablation study of Relation LoRA position.
    }
    \vspace{-1mm}
    \label{tab:ablation_relationLoRA_pos}
\end{table}
To determine the optimal position of Relation LoRAs, we experiment with different settings in the query (Q), key (K), and value (V) matrices, as shown in Tab.~\ref{tab:ablation_relationLoRA_pos}.
Inserting Relation LoRAs to the V matrix results in the lowest Relation Accuracy,
likely because V features predominantly exhibit appearance information, making it challenging to accurately capture the desired relations.
In contrast, placing Relation LoRAs in the Q matrix or KV matrices is suboptimal 
since the overlapping nature of the QK matrices hinders their ability to process different information separately, which is not conducive to decoupling relations from appearances.
In contrast, inserting Relation LoRAs to the QK matrices achieves the best Relation Accuracy, consistent with our analysis of full attention in Fig.~\ref{fig:model_analysis}.

\noindent\textbf{Ablation on \CLName (\CLNameShort).}\quad
\begin{table}[t]
    \centering
    \vspace{-1.5mm}
    \resizebox{\columnwidth}{!}{
    \begin{tabular}{lcccc} 
        \textbf{Method} & \tabincell{c}{\textbf{Relation}\\\textbf{Accuracy}} & 
        \textbf{CLIP-T} & \tabincell{c}{\textbf{Temporal}\\\textbf{Consistency}} & 
        \textbf{FVD$\downarrow$} \\ 
        \shline
        \noalign{\smallskip}
        MotionInversion~\cite{wang2024motioninversion} & 0.3151\std{$\pm$0.03} & 0.3217 & 0.9855 & 2084.51 \\
         MotionInversion + \CLNameShort & 0.3633\std{$\pm$0.05} & 0.3181 & 0.9862 & 2063.30
    \end{tabular}
    }
    \caption{Effects of \CLName on motion customization method (MotionInversion).
    }
    \vspace{-4mm}
    \label{tab:ablation_CL_on_MI}
\end{table}
To verify the effectiveness of \CLNameShort among different methods, we integrate it with MotionInversion~\cite{wang2024motioninversion}.
Results in Tab.~\ref{tab:ablation_CL_on_MI} show that incorporating \CLNameShort enhances Relation Accuracy and Temporal Consistency while maintaining comparable CLIP-T, demonstrating its potential for generalization across different methods.

\section{Conclusion}
\label{sec:conclusion}
In this paper, we present \frameworkplain, a novel relational video customization method that 
accurately models complex relations defined in exemplar videos through relational decoupling learning and relational dynamics enhancement. 
We introduce \tripletLoRAName to decompose relations into appearance and relational information and further enhance this decoupling with \trainStrategyName.
Our analysis of query, key, and value features in MM-DiT's full attention motivates and offers interpretability for our model design.
To further enhance relation dynamics learning, we propose \CLName, which prioritizes relational dynamics over detailed appearances.
Extensive experimental results demonstrate the superior customization capabilities of \frameworkplain.
\\
\textbf{Limitations.}\quad
Existing metrics for relation accuracy may not fully capture the customization capabilities of models.
While the use of VLMs simplifies evaluation and reduces bias, the metric relies on VLM's capabilities; future work should develop metrics that align better with human perception.

% \clearpage

{
    \small
    \bibliographystyle{ieeenat_fullname}
    \bibliography{main}

\begin{thebibliography}{103}
\providecommand{\natexlab}[1]{#1}
\providecommand{\url}[1]{\texttt{#1}}
\expandafter\ifx\csname urlstyle\endcsname\relax
  \providecommand{\doi}[1]{doi: #1}\else
  \providecommand{\doi}{doi: \begingroup \urlstyle{rm}\Url}\fi

\bibitem[An et~al.(2023)An, Zhang, Yang, Gupta, Huang, Luo, and Yin]{latent_shift}
Jie An, Songyang Zhang, Harry Yang, Sonal Gupta, Jia-Bin Huang, Jiebo Luo, and Xi Yin.
\newblock Latent-shift: Latent diffusion with temporal shift for efficient text-to-video generation.
\newblock \emph{arXiv preprint arXiv:2304.08477}, 2023.

\bibitem[Bai et~al.(2023)Bai, Bai, Yang, Wang, Tan, Wang, Lin, Zhou, and Zhou]{Qwen-VL}
Jinze Bai, Shuai Bai, Shusheng Yang, Shijie Wang, Sinan Tan, Peng Wang, Junyang Lin, Chang Zhou, and Jingren Zhou.
\newblock Qwen-vl: A versatile vision-language model for understanding, localization, text reading, and beyond.
\newblock \emph{arXiv preprint arXiv:2308.12966}, 2023.

\bibitem[Bar-Tal et~al.(2024)Bar-Tal, Chefer, Tov, Herrmann, Paiss, Zada, Ephrat, Hur, Li, Michaeli, et~al.]{bar2024lumiere}
Omer Bar-Tal, Hila Chefer, Omer Tov, Charles Herrmann, Roni Paiss, Shiran Zada, Ariel Ephrat, Junhwa Hur, Yuanzhen Li, Tomer Michaeli, et~al.
\newblock Lumiere: A space-time diffusion model for video generation.
\newblock \emph{arXiv preprint arXiv:2401.12945}, 2024.

\bibitem[Blattmann et~al.(2023)Blattmann, Dockhorn, Kulal, Mendelevitch, Kilian, Lorenz, Levi, English, Voleti, Letts, et~al.]{svd}
Andreas Blattmann, Tim Dockhorn, Sumith Kulal, Daniel Mendelevitch, Maciej Kilian, Dominik Lorenz, Yam Levi, Zion English, Vikram Voleti, Adam Letts, et~al.
\newblock Stable video diffusion: Scaling latent video diffusion models to large datasets.
\newblock \emph{arXiv preprint arXiv:2311.15127}, 2023.

\bibitem[Brooks et~al.(2024)Brooks, Peebles, Holmes, DePue, Guo, Jing, Schnurr, Taylor, Luhman, Luhman, Ng, Wang, and Ramesh]{sora}
Tim Brooks, Bill Peebles, Connor Holmes, Will DePue, Yufei Guo, Li Jing, David Schnurr, Joe Taylor, Troy Luhman, Eric Luhman, Clarence Ng, Ricky Wang, and Aditya Ramesh.
\newblock Video generation models as world simulators.
\newblock 2024.

\bibitem[Cai et~al.(2024)Cai, Cun, Li, Liu, Zhang, Zhang, Shan, and Yue]{cai2024ditctrl}
Minghong Cai, Xiaodong Cun, Xiaoyu Li, Wenze Liu, Zhaoyang Zhang, Yong Zhang, Ying Shan, and Xiangyu Yue.
\newblock Ditctrl: Exploring attention control in multi-modal diffusion transformer for tuning-free multi-prompt longer video generation.
\newblock \emph{arXiv preprint arXiv:2412.18597}, 2024.

\bibitem[Cao et~al.(2023)Cao, Wang, Qi, Shan, Qie, and Zheng]{cao2023masactrl}
Mingdeng Cao, Xintao Wang, Zhongang Qi, Ying Shan, Xiaohu Qie, and Yinqiang Zheng.
\newblock Masactrl: Tuning-free mutual self-attention control for consistent image synthesis and editing.
\newblock In \emph{Proceedings of the IEEE/CVF international conference on computer vision}, pages 22560--22570, 2023.

\bibitem[Chefer et~al.(2024)Chefer, Zada, Paiss, Ephrat, Tov, Rubinstein, Wolf, Dekel, Michaeli, and Mosseri]{chefer2024still_moving}
Hila Chefer, Shiran Zada, Roni Paiss, Ariel Ephrat, Omer Tov, Michael Rubinstein, Lior Wolf, Tali Dekel, Tomer Michaeli, and Inbar Mosseri.
\newblock Still-moving: Customized video generation without customized video data.
\newblock \emph{arXiv preprint arXiv:2407.08674}, 2024.

\bibitem[Chen et~al.(2023{\natexlab{a}})Chen, Wang, Zeng, Zhang, Zhou, Han, and Zhu]{chen2023videodreamer}
Hong Chen, Xin Wang, Guanning Zeng, Yipeng Zhang, Yuwei Zhou, Feilin Han, and Wenwu Zhu.
\newblock Videodreamer: Customized multi-subject text-to-video generation with disen-mix finetuning.
\newblock \emph{arXiv preprint arXiv:2311.00990}, 2023{\natexlab{a}}.

\bibitem[Chen et~al.(2023{\natexlab{b}})Chen, Xia, He, Zhang, Cun, Yang, Xing, Liu, Chen, Wang, et~al.]{chen2023videocrafter1}
Haoxin Chen, Menghan Xia, Yingqing He, Yong Zhang, Xiaodong Cun, Shaoshu Yang, Jinbo Xing, Yaofang Liu, Qifeng Chen, Xintao Wang, et~al.
\newblock Videocrafter1: Open diffusion models for high-quality video generation.
\newblock \emph{arXiv preprint arXiv:2310.19512}, 2023{\natexlab{b}}.

\bibitem[Chen et~al.(2024{\natexlab{a}})Chen, Wang, Zhang, Zhou, Zhang, Tang, and Zhu]{chen2024disenstudio}
Hong Chen, Xin Wang, Yipeng Zhang, Yuwei Zhou, Zeyang Zhang, Siao Tang, and Wenwu Zhu.
\newblock Disenstudio: Customized multi-subject text-to-video generation with disentangled spatial control.
\newblock \emph{arXiv preprint arXiv:2405.12796}, 2024{\natexlab{a}}.

\bibitem[Chen et~al.(2024{\natexlab{b}})Chen, Zhang, Cun, Xia, Wang, Weng, and Shan]{chen2024videocrafter2}
Haoxin Chen, Yong Zhang, Xiaodong Cun, Menghan Xia, Xintao Wang, Chao Weng, and Ying Shan.
\newblock Videocrafter2: Overcoming data limitations for high-quality video diffusion models.
\newblock In \emph{Proceedings of the IEEE/CVF Conference on Computer Vision and Pattern Recognition}, pages 7310--7320, 2024{\natexlab{b}}.

\bibitem[Chen et~al.(2025)Chen, Siarohin, Menapace, Fang, Lee, Skorokhodov, Aberman, Zhu, Yang, and Tulyakov]{chen2025multi}
Tsai-Shien Chen, Aliaksandr Siarohin, Willi Menapace, Yuwei Fang, Kwot~Sin Lee, Ivan Skorokhodov, Kfir Aberman, Jun-Yan Zhu, Ming-Hsuan Yang, and Sergey Tulyakov.
\newblock Multi-subject open-set personalization in video generation.
\newblock \emph{arXiv preprint arXiv:2501.06187}, 2025.

\bibitem[Chen et~al.(2024{\natexlab{c}})Chen, Hu, Li, Ruiz, Jia, Chang, and Cohen]{SuTI}
Wenhu Chen, Hexiang Hu, Yandong Li, Nataniel Ruiz, Xuhui Jia, Ming-Wei Chang, and William~W Cohen.
\newblock Subject-driven text-to-image generation via apprenticeship learning.
\newblock \emph{Advances in Neural Information Processing Systems}, 36, 2024{\natexlab{c}}.

\bibitem[Dalva and Yanardag(2024)]{dalva2024noiseclr}
Yusuf Dalva and Pinar Yanardag.
\newblock Noiseclr: A contrastive learning approach for unsupervised discovery of interpretable directions in diffusion models.
\newblock In \emph{Proceedings of the IEEE/CVF Conference on Computer Vision and Pattern Recognition}, pages 24209--24218, 2024.

\bibitem[Esser et~al.(2023)Esser, Chiu, Atighehchian, Granskog, and Germanidis]{esser2023structure}
Patrick Esser, Johnathan Chiu, Parmida Atighehchian, Jonathan Granskog, and Anastasis Germanidis.
\newblock Structure and content-guided video synthesis with diffusion models.
\newblock In \emph{Proceedings of the IEEE/CVF International Conference on Computer Vision}, pages 7346--7356, 2023.

\bibitem[Esser et~al.(2024)Esser, Kulal, Blattmann, Entezari, M{\"u}ller, Saini, Levi, Lorenz, Sauer, Boesel, et~al.]{SD3}
Patrick Esser, Sumith Kulal, Andreas Blattmann, Rahim Entezari, Jonas M{\"u}ller, Harry Saini, Yam Levi, Dominik Lorenz, Axel Sauer, Frederic Boesel, et~al.
\newblock Scaling rectified flow transformers for high-resolution image synthesis.
\newblock In \emph{Forty-first International Conference on Machine Learning}, 2024.

\bibitem[Fan et~al.(2025)Fan, Si, Song, Yang, He, Zhuo, Huang, Dong, He, Pan, et~al.]{Vchitect}
Weichen Fan, Chenyang Si, Junhao Song, Zhenyu Yang, Yinan He, Long Zhuo, Ziqi Huang, Ziyue Dong, Jingwen He, Dongwei Pan, et~al.
\newblock Vchitect-2.0: Parallel transformer for scaling up video diffusion models.
\newblock \emph{arXiv preprint arXiv:2501.08453}, 2025.

\bibitem[Gal et~al.(2022)Gal, Alaluf, Atzmon, Patashnik, Bermano, Chechik, and Cohen-Or]{textInversion}
Rinon Gal, Yuval Alaluf, Yuval Atzmon, Or Patashnik, Amit~H Bermano, Gal Chechik, and Daniel Cohen-Or.
\newblock An image is worth one word: Personalizing text-to-image generation using textual inversion.
\newblock \emph{arXiv preprint arXiv:2208.01618}, 2022.

\bibitem[Gao et~al.(2020)Gao, Liu, Zhu, Liu, Cao, He, He, and Yan]{gao2020interactgan}
Chen Gao, Si Liu, Defa Zhu, Quan Liu, Jie Cao, Haoqian He, Ran He, and Shuicheng Yan.
\newblock Interactgan: Learning to generate human-object interaction.
\newblock In \emph{Proceedings of the 28th ACM International Conference on Multimedia}, pages 165--173, 2020.

\bibitem[Ge et~al.(2024)Ge, Jia, Isobe, Li, Wang, Mu, Zhou, Wang, Lu, Tian, et~al.]{ge2024customizing}
Mengmeng Ge, Xu Jia, Takashi Isobe, Xiaomin Li, Qinghe Wang, Jing Mu, Dong Zhou, Li Wang, Huchuan Lu, Lu Tian, et~al.
\newblock Customizing text-to-image generation with inverted interaction.
\newblock In \emph{Proceedings of the 32nd ACM International Conference on Multimedia}, pages 10901--10909, 2024.

\bibitem[Guo et~al.(2023)Guo, Yang, Rao, Wang, Qiao, Lin, and Dai]{animatediff}
Yuwei Guo, Ceyuan Yang, Anyi Rao, Yaohui Wang, Yu Qiao, Dahua Lin, and Bo Dai.
\newblock Animatediff: Animate your personalized text-to-image diffusion models without specific tuning.
\newblock \emph{arXiv preprint arXiv:2307.04725}, 2023.

\bibitem[Hamm and Lee(2008)]{hamm2008grassmann}
Jihun Hamm and Daniel~D Lee.
\newblock Grassmann discriminant analysis: a unifying view on subspace-based learning.
\newblock In \emph{Proceedings of the 25th international conference on Machine learning}, pages 376--383, 2008.

\bibitem[He et~al.(2024)He, Liu, Qian, Wang, Hu, Cao, Yan, Zhou, and Zhang]{he2024id_animator}
Xuanhua He, Quande Liu, Shengju Qian, Xin Wang, Tao Hu, Ke Cao, Keyu Yan, Man Zhou, and Jie Zhang.
\newblock Id-animator: Zero-shot identity-preserving human video generation.
\newblock \emph{arXiv preprint arXiv:2404.15275}, 2024.

\bibitem[Ho and Salimans(2022)]{ho2022classifier_free_guide}
Jonathan Ho and Tim Salimans.
\newblock Classifier-free diffusion guidance.
\newblock \emph{arXiv preprint arXiv:2207.12598}, 2022.

\bibitem[Ho et~al.(2020)Ho, Jain, and Abbeel]{DDPM}
Jonathan Ho, Ajay Jain, and Pieter Abbeel.
\newblock Denoising diffusion probabilistic models.
\newblock \emph{Advances in neural information processing systems}, 33:\penalty0 6840--6851, 2020.

\bibitem[Ho et~al.(2022{\natexlab{a}})Ho, Chan, Saharia, Whang, Gao, Gritsenko, Kingma, Poole, Norouzi, Fleet, et~al.]{imagenVideo}
Jonathan Ho, William Chan, Chitwan Saharia, Jay Whang, Ruiqi Gao, Alexey Gritsenko, Diederik~P Kingma, Ben Poole, Mohammad Norouzi, David~J Fleet, et~al.
\newblock Imagen video: High definition video generation with diffusion models.
\newblock \emph{arXiv preprint arXiv:2210.02303}, 2022{\natexlab{a}}.

\bibitem[Ho et~al.(2022{\natexlab{b}})Ho, Salimans, Gritsenko, Chan, Norouzi, and Fleet]{VDM}
Jonathan Ho, Tim Salimans, Alexey Gritsenko, William Chan, Mohammad Norouzi, and David~J. Fleet.
\newblock Video diffusion models.
\newblock \emph{arXiv preprint arXiv:2204.03458}, 2022{\natexlab{b}}.

\bibitem[Hoe et~al.(2024)Hoe, Jiang, Chan, Tan, and Hu]{hoe2024interactdiffusion}
Jiun~Tian Hoe, Xudong Jiang, Chee~Seng Chan, Yap-Peng Tan, and Weipeng Hu.
\newblock Interactdiffusion: Interaction control in text-to-image diffusion models.
\newblock In \emph{Proceedings of the IEEE/CVF Conference on Computer Vision and Pattern Recognition}, pages 6180--6189, 2024.

\bibitem[Hu et~al.(2021)Hu, Shen, Wallis, Allen-Zhu, Li, Wang, Wang, and Chen]{hu2021lora}
Edward~J Hu, Yelong Shen, Phillip Wallis, Zeyuan Allen-Zhu, Yuanzhi Li, Shean Wang, Lu Wang, and Weizhu Chen.
\newblock Lora: Low-rank adaptation of large language models.
\newblock \emph{arXiv preprint arXiv:2106.09685}, 2021.

\bibitem[Hua et~al.(2021)Hua, Zheng, Bai, Zhang, Zhang, and Mei]{hua2021exploiting}
Tianyu Hua, Hongdong Zheng, Yalong Bai, Wei Zhang, Xiao-Ping Zhang, and Tao Mei.
\newblock Exploiting relationship for complex-scene image generation.
\newblock In \emph{Proceedings of the AAAI Conference on Artificial Intelligence}, pages 1584--1592, 2021.

\bibitem[Huang et~al.(2025)Huang, Yuan, Liu, Wang, Wang, Zhang, Wan, Zhang, and Gai]{huang2025conceptmaster}
Yuzhou Huang, Ziyang Yuan, Quande Liu, Qiulin Wang, Xintao Wang, Ruimao Zhang, Pengfei Wan, Di Zhang, and Kun Gai.
\newblock Conceptmaster: Multi-concept video customization on diffusion transformer models without test-time tuning.
\newblock \emph{arXiv preprint arXiv:2501.04698}, 2025.

\bibitem[Huang et~al.(2024)Huang, Wu, Jiang, Chan, and Liu]{huang2024reversion}
Ziqi Huang, Tianxing Wu, Yuming Jiang, Kelvin~CK Chan, and Ziwei Liu.
\newblock Reversion: Diffusion-based relation inversion from images.
\newblock In \emph{SIGGRAPH Asia 2024 Conference Papers}, pages 1--11, 2024.

\bibitem[Jeong et~al.(2024{\natexlab{a}})Jeong, Chang, Park, and Ye]{jeong2024dreammotion}
Hyeonho Jeong, Jinho Chang, Geon~Yeong Park, and Jong~Chul Ye.
\newblock Dreammotion: Space-time self-similar score distillation for zero-shot video editing.
\newblock In \emph{European Conference on Computer Vision}, pages 358--376. Springer, 2024{\natexlab{a}}.

\bibitem[Jeong et~al.(2024{\natexlab{b}})Jeong, Park, and Ye]{jeong2024vmc}
Hyeonho Jeong, Geon~Yeong Park, and Jong~Chul Ye.
\newblock Vmc: Video motion customization using temporal attention adaption for text-to-video diffusion models.
\newblock In \emph{Proceedings of the IEEE/CVF Conference on Computer Vision and Pattern Recognition}, pages 9212--9221, 2024{\natexlab{b}}.

\bibitem[Kingma and Welling(2013)]{vae}
Diederik~P Kingma and Max Welling.
\newblock Auto-encoding variational bayes.
\newblock \emph{arXiv preprint arXiv:1312.6114}, 2013.

\bibitem[Kirillov et~al.(2023)Kirillov, Mintun, Ravi, Mao, Rolland, Gustafson, Xiao, Whitehead, Berg, Lo, et~al.]{sam}
Alexander Kirillov, Eric Mintun, Nikhila Ravi, Hanzi Mao, Chloe Rolland, Laura Gustafson, Tete Xiao, Spencer Whitehead, Alexander~C Berg, Wan-Yen Lo, et~al.
\newblock Segment anything.
\newblock In \emph{Proceedings of the IEEE/CVF International Conference on Computer Vision}, pages 4015--4026, 2023.

\bibitem[Kondratyuk et~al.(2023)Kondratyuk, Yu, Gu, Lezama, Huang, Hornung, Adam, Akbari, Alon, Birodkar, et~al.]{kondratyuk2023videopoet}
Dan Kondratyuk, Lijun Yu, Xiuye Gu, Jos{\'e} Lezama, Jonathan Huang, Rachel Hornung, Hartwig Adam, Hassan Akbari, Yair Alon, Vighnesh Birodkar, et~al.
\newblock Videopoet: A large language model for zero-shot video generation.
\newblock \emph{arXiv preprint arXiv:2312.14125}, 2023.

\bibitem[Kong et~al.(2024)Kong, Tian, Zhang, Min, Dai, Zhou, Xiong, Li, Wu, Zhang, et~al.]{kong2024hunyuanvideo}
Weijie Kong, Qi Tian, Zijian Zhang, Rox Min, Zuozhuo Dai, Jin Zhou, Jiangfeng Xiong, Xin Li, Bo Wu, Jianwei Zhang, et~al.
\newblock Hunyuanvideo: A systematic framework for large video generative models.
\newblock \emph{arXiv preprint arXiv:2412.03603}, 2024.

\bibitem[Lab and etc.(2024)]{Open-Sora-Plan}
PKU-Yuan Lab and Tuzhan~AI etc.
\newblock Open-sora: Democratizing efficient video production for all, 2024.
\newblock https://doi.org/10. 5281/zenodo.10948109.

\bibitem[Li et~al.(2024)Li, Qiu, Zhang, Wang, Wei, Li, Zhang, Wu, and Cai]{li2024personalvideo}
Hengjia Li, Haonan Qiu, Shiwei Zhang, Xiang Wang, Yujie Wei, Zekun Li, Yingya Zhang, Boxi Wu, and Deng Cai.
\newblock Personalvideo: High id-fidelity video customization without dynamic and semantic degradation.
\newblock \emph{arXiv preprint arXiv:2411.17048}, 2024.

\bibitem[Lipman et~al.(2022)Lipman, Chen, Ben-Hamu, Nickel, and Le]{lipman2022flow}
Yaron Lipman, Ricky~TQ Chen, Heli Ben-Hamu, Maximilian Nickel, and Matt Le.
\newblock Flow matching for generative modeling.
\newblock \emph{arXiv preprint arXiv:2210.02747}, 2022.

\bibitem[Liu et~al.(2024{\natexlab{a}})Liu, Zhang, Wang, Wei, Qiu, Zhao, Zhang, Ye, and Wan]{liu2024timestep}
Feng Liu, Shiwei Zhang, Xiaofeng Wang, Yujie Wei, Haonan Qiu, Yuzhong Zhao, Yingya Zhang, Qixiang Ye, and Fang Wan.
\newblock Timestep embedding tells: It's time to cache for video diffusion model.
\newblock \emph{arXiv preprint arXiv:2411.19108}, 2024{\natexlab{a}}.

\bibitem[Liu et~al.(2019)Liu, Shahroudy, Perez, Wang, Duan, and Kot]{liu2019ntu}
Jun Liu, Amir Shahroudy, Mauricio Perez, Gang Wang, Ling-Yu Duan, and Alex~C Kot.
\newblock Ntu rgb+ d 120: A large-scale benchmark for 3d human activity understanding.
\newblock \emph{IEEE transactions on pattern analysis and machine intelligence}, 42\penalty0 (10):\penalty0 2684--2701, 2019.

\bibitem[Liu et~al.(2023)Liu, Zeng, Ren, Li, Zhang, Yang, Li, Yang, Su, Zhu, et~al.]{groundingdino}
Shilong Liu, Zhaoyang Zeng, Tianhe Ren, Feng Li, Hao Zhang, Jie Yang, Chunyuan Li, Jianwei Yang, Hang Su, Jun Zhu, et~al.
\newblock Grounding dino: Marrying dino with grounded pre-training for open-set object detection.
\newblock \emph{arXiv preprint arXiv:2303.05499}, 2023.

\bibitem[Liu et~al.(2022)Liu, Gong, and Liu]{liu2022flow}
Xingchao Liu, Chengyue Gong, and Qiang Liu.
\newblock Flow straight and fast: Learning to generate and transfer data with rectified flow.
\newblock \emph{arXiv preprint arXiv:2209.03003}, 2022.

\bibitem[Liu et~al.(2024{\natexlab{b}})Liu, Li, Xie, Li, Ge, Liu, and Jin]{liu2024towards}
Zhihang Liu, Jun Li, Hongtao Xie, Pandeng Li, Jiannan Ge, Sun-Ao Liu, and Guoqing Jin.
\newblock Towards balanced alignment: Modal-enhanced semantic modeling for video moment retrieval.
\newblock In \emph{Proceedings of the AAAI conference on artificial intelligence}, pages 3855--3863, 2024{\natexlab{b}}.

\bibitem[Loshchilov(2017)]{adamw}
I Loshchilov.
\newblock Decoupled weight decay regularization.
\newblock \emph{arXiv preprint arXiv:1711.05101}, 2017.

\bibitem[Ma et~al.(2024{\natexlab{a}})Ma, Wang, Jia, Chen, Liu, Li, Chen, and Qiao]{ma2024latte}
Xin Ma, Yaohui Wang, Gengyun Jia, Xinyuan Chen, Ziwei Liu, Yuan-Fang Li, Cunjian Chen, and Yu Qiao.
\newblock Latte: Latent diffusion transformer for video generation.
\newblock \emph{arXiv preprint arXiv:2401.03048}, 2024{\natexlab{a}}.

\bibitem[Ma et~al.(2024{\natexlab{b}})Ma, Zhou, Yeh, Wang, Li, Yang, Dong, Keutzer, and Feng]{ma2024magicme}
Ze Ma, Daquan Zhou, Chun-Hsiao Yeh, Xue-She Wang, Xiuyu Li, Huanrui Yang, Zhen Dong, Kurt Keutzer, and Jiashi Feng.
\newblock Magic-me: Identity-specific video customized diffusion.
\newblock \emph{arXiv preprint arXiv:2402.09368}, 2024{\natexlab{b}}.

\bibitem[Miech et~al.(2020)Miech, Alayrac, Smaira, Laptev, Sivic, and Zisserman]{miech2020end}
Antoine Miech, Jean-Baptiste Alayrac, Lucas Smaira, Ivan Laptev, Josef Sivic, and Andrew Zisserman.
\newblock End-to-end learning of visual representations from uncurated instructional videos.
\newblock In \emph{Proceedings of the IEEE/CVF conference on computer vision and pattern recognition}, pages 9879--9889, 2020.

\bibitem[Molad et~al.(2023)Molad, Horwitz, Valevski, Acha, Matias, Pritch, Leviathan, and Hoshen]{dreamix}
Eyal Molad, Eliahu Horwitz, Dani Valevski, Alex~Rav Acha, Yossi Matias, Yael Pritch, Yaniv Leviathan, and Yedid Hoshen.
\newblock Dreamix: Video diffusion models are general video editors.
\newblock \emph{arXiv preprint arXiv:2302.01329}, 2023.

\bibitem[Ng et~al.(2022)Ng, Ong, Zheng, Ni, Yeo, and Liu]{Animal_Kingdom}
Xun~Long Ng, Kian~Eng Ong, Qichen Zheng, Yun Ni, Si~Yong Yeo, and Jun Liu.
\newblock Animal kingdom: A large and diverse dataset for animal behavior understanding.
\newblock In \emph{Proceedings of the IEEE/CVF Conference on Computer Vision and Pattern Recognition (CVPR)}, pages 19023--19034, 2022.

\bibitem[Oord et~al.(2018)Oord, Li, and Vinyals]{oord2018representation}
Aaron van~den Oord, Yazhe Li, and Oriol Vinyals.
\newblock Representation learning with contrastive predictive coding.
\newblock \emph{arXiv preprint arXiv:1807.03748}, 2018.

\bibitem[Peebles and Xie(2023)]{DiT}
William Peebles and Saining Xie.
\newblock Scalable diffusion models with transformers.
\newblock In \emph{Proceedings of the IEEE/CVF International Conference on Computer Vision}, pages 4195--4205, 2023.

\bibitem[Podell et~al.(2023)Podell, English, Lacey, Blattmann, Dockhorn, M{\"u}ller, Penna, and Rombach]{podell2023sdxl}
Dustin Podell, Zion English, Kyle Lacey, Andreas Blattmann, Tim Dockhorn, Jonas M{\"u}ller, Joe Penna, and Robin Rombach.
\newblock Sdxl: Improving latent diffusion models for high-resolution image synthesis.
\newblock \emph{arXiv preprint arXiv:2307.01952}, 2023.

\bibitem[Qing et~al.(2024)Qing, Zhang, Wang, Wang, Wei, Zhang, Gao, and Sang]{qing2024hierarchical}
Zhiwu Qing, Shiwei Zhang, Jiayu Wang, Xiang Wang, Yujie Wei, Yingya Zhang, Changxin Gao, and Nong Sang.
\newblock Hierarchical spatio-temporal decoupling for text-to-video generation.
\newblock In \emph{Proceedings of the IEEE/CVF Conference on Computer Vision and Pattern Recognition}, pages 6635--6645, 2024.

\bibitem[Ren et~al.(2024)Ren, Zhou, Yang, Shi, Liu, Liu, Kwon, and Shrivastava]{customize_a_video}
Yixuan Ren, Yang Zhou, Jimei Yang, Jing Shi, Difan Liu, Feng Liu, Mingi Kwon, and Abhinav Shrivastava.
\newblock Customize-a-video: One-shot motion customization of text-to-video diffusion models.
\newblock \emph{arXiv preprint arXiv:2402.14780}, 2024.

\bibitem[Rombach et~al.(2022)Rombach, Blattmann, Lorenz, Esser, and Ommer]{stableDiffusion}
Robin Rombach, Andreas Blattmann, Dominik Lorenz, Patrick Esser, and Bj{\"o}rn Ommer.
\newblock High-resolution image synthesis with latent diffusion models.
\newblock In \emph{Proceedings of the IEEE/CVF Conference on Computer Vision and Pattern Recognition}, pages 10684--10695, 2022.

\bibitem[Ruiz et~al.(2023)Ruiz, Li, Jampani, Pritch, Rubinstein, and Aberman]{dreambooth}
Nataniel Ruiz, Yuanzhen Li, Varun Jampani, Yael Pritch, Michael Rubinstein, and Kfir Aberman.
\newblock Dreambooth: Fine tuning text-to-image diffusion models for subject-driven generation.
\newblock In \emph{Proceedings of the IEEE/CVF Conference on Computer Vision and Pattern Recognition}, pages 22500--22510, 2023.

\bibitem[Shahroudy et~al.(2016)Shahroudy, Liu, Ng, and Wang]{shahroudy2016ntu}
Amir Shahroudy, Jun Liu, Tian-Tsong Ng, and Gang Wang.
\newblock Ntu rgb+ d: A large scale dataset for 3d human activity analysis.
\newblock In \emph{Proceedings of the IEEE conference on computer vision and pattern recognition}, pages 1010--1019, 2016.

\bibitem[She et~al.(2025)She, Liu, Pang, Wang, Yang, He, Zhang, Wang, Huang, Tang, et~al.]{she2025customvideox}
D She, Mushui Liu, Jingxuan Pang, Jin Wang, Zhen Yang, Wanggui He, Guanghao Zhang, Yi Wang, Qihan Huang, Haobin Tang, et~al.
\newblock Customvideox: 3d reference attention driven dynamic adaptation for zero-shot customized video diffusion transformers.
\newblock \emph{arXiv preprint arXiv:2502.06527}, 2025.

\bibitem[Shi et~al.(2024)Shi, Qi, Wu, Bai, Wang, Tong, Li, and Yang]{shi2024relationbooth}
Qingyu Shi, Lu Qi, Jianzong Wu, Jinbin Bai, Jingbo Wang, Yunhai Tong, Xiangtai Li, and Ming-Husan Yang.
\newblock Relationbooth: Towards relation-aware customized object generation.
\newblock \emph{arXiv preprint arXiv:2410.23280}, 2024.

\bibitem[Tan et~al.(2024{\natexlab{a}})Tan, Gong, Feng, Zheng, Zheng, Shi, Shen, Chen, and Yang]{tan2024mimir}
Shuai Tan, Biao Gong, Yutong Feng, Kecheng Zheng, Dandan Zheng, Shuwei Shi, Yujun Shen, Jingdong Chen, and Ming Yang.
\newblock Mimir: Improving video diffusion models for precise text understanding.
\newblock \emph{arXiv preprint arXiv:2412.03085}, 2024{\natexlab{a}}.

\bibitem[Tan et~al.(2024{\natexlab{b}})Tan, Gong, Wang, Zhang, Zheng, Zheng, Zheng, Chen, and Yang]{tan2024animate}
Shuai Tan, Biao Gong, Xiang Wang, Shiwei Zhang, Dandan Zheng, Ruobing Zheng, Kecheng Zheng, Jingdong Chen, and Ming Yang.
\newblock Animate-x: Universal character image animation with enhanced motion representation.
\newblock \emph{arXiv preprint arXiv:2410.10306}, 2024{\natexlab{b}}.

\bibitem[Tan et~al.(2024{\natexlab{c}})Tan, Ji, Bi, and Pan]{tan2025edtalk}
Shuai Tan, Bin Ji, Mengxiao Bi, and Ye Pan.
\newblock Edtalk: Efficient disentanglement for emotional talking head synthesis.
\newblock In \emph{European Conference on Computer Vision}, pages 398--416. Springer, 2024{\natexlab{c}}.

\bibitem[Tan et~al.(2024{\natexlab{d}})Tan, Ji, and Pan]{tan2024flowvqtalker}
Shuai Tan, Bin Ji, and Ye Pan.
\newblock Flowvqtalker: High-quality emotional talking face generation through normalizing flow and quantization.
\newblock In \emph{Proceedings of the IEEE/CVF Conference on Computer Vision and Pattern Recognition}, pages 26317--26327, 2024{\natexlab{d}}.

\bibitem[Tang et~al.(2020)Tang, Niu, Huang, Shi, and Zhang]{tang2020unbiased-SGG}
Kaihua Tang, Yulei Niu, Jianqiang Huang, Jiaxin Shi, and Hanwang Zhang.
\newblock Unbiased scene graph generation from biased training.
\newblock In \emph{Proceedings of the IEEE/CVF conference on computer vision and pattern recognition}, pages 3716--3725, 2020.

\bibitem[Team(2024)]{genmo2024mochi}
Genmo Team.
\newblock Mochi 1.
\newblock \url{https://github.com/genmoai/models}, 2024.

\bibitem[Tu et~al.(2024{\natexlab{a}})Tu, Dai, Cheng, Hu, Han, Wu, and Jiang]{tu2024motioneditor}
Shuyuan Tu, Qi Dai, Zhi-Qi Cheng, Han Hu, Xintong Han, Zuxuan Wu, and Yu-Gang Jiang.
\newblock Motioneditor: Editing video motion via content-aware diffusion.
\newblock In \emph{Proceedings of the IEEE/CVF Conference on Computer Vision and Pattern Recognition}, pages 7882--7891, 2024{\natexlab{a}}.

\bibitem[Tu et~al.(2024{\natexlab{b}})Tu, Dai, Zhang, Xie, Cheng, Luo, Han, Wu, and Jiang]{tu2024motionfollower}
Shuyuan Tu, Qi Dai, Zihao Zhang, Sicheng Xie, Zhi-Qi Cheng, Chong Luo, Xintong Han, Zuxuan Wu, and Yu-Gang Jiang.
\newblock Motionfollower: Editing video motion via lightweight score-guided diffusion.
\newblock \emph{arXiv preprint arXiv:2405.20325}, 2024{\natexlab{b}}.

\bibitem[Vaswani et~al.(2017)Vaswani, Shazeer, Parmar, Uszkoreit, Jones, Gomez, Kaiser, and Polosukhin]{vaswani2017attention}
Ashish Vaswani, Noam Shazeer, Niki Parmar, Jakob Uszkoreit, Llion Jones, Aidan~N Gomez, {\L}ukasz Kaiser, and Illia Polosukhin.
\newblock Attention is all you need.
\newblock \emph{Advances in neural information processing systems}, 30, 2017.

\bibitem[Wang et~al.(2023{\natexlab{a}})Wang, Yuan, Chen, Zhang, Wang, and Zhang]{modelScope}
Jiuniu Wang, Hangjie Yuan, Dayou Chen, Yingya Zhang, Xiang Wang, and Shiwei Zhang.
\newblock Modelscope text-to-video technical report.
\newblock \emph{arXiv preprint arXiv:2308.06571}, 2023{\natexlab{a}}.

\bibitem[Wang et~al.(2024{\natexlab{a}})Wang, Mai, Shen, Liang, Tao, Wan, Zhang, Li, and Chen]{wang2024motioninversion}
Luozhou Wang, Ziyang Mai, Guibao Shen, Yixun Liang, Xin Tao, Pengfei Wan, Di Zhang, Yijun Li, and Yingcong Chen.
\newblock Motion inversion for video customization.
\newblock \emph{arXiv preprint arXiv:2403.20193}, 2024{\natexlab{a}}.

\bibitem[Wang et~al.(2023{\natexlab{b}})Wang, Yuan, Zhang, Chen, Wang, Zhang, Shen, Zhao, and Zhou]{wang2023videocomposer}
Xiang Wang, Hangjie Yuan, Shiwei Zhang, Dayou Chen, Jiuniu Wang, Yingya Zhang, Yujun Shen, Deli Zhao, and Jingren Zhou.
\newblock Videocomposer: Compositional video synthesis with motion controllability.
\newblock \emph{Advances in Neural Information Processing Systems}, 36:\penalty0 7594--7611, 2023{\natexlab{b}}.

\bibitem[Wang et~al.(2023{\natexlab{c}})Wang, Zhang, Zhang, Liu, Zhang, Gao, and Sang]{wang2023videolcm}
Xiang Wang, Shiwei Zhang, Han Zhang, Yu Liu, Yingya Zhang, Changxin Gao, and Nong Sang.
\newblock Videolcm: Video latent consistency model.
\newblock \emph{arXiv preprint arXiv:2312.09109}, 2023{\natexlab{c}}.

\bibitem[Wang et~al.(2024{\natexlab{b}})Wang, Zhang, Gao, Wang, Zhou, Zhang, Yan, and Sang]{wang2024unianimate}
Xiang Wang, Shiwei Zhang, Changxin Gao, Jiayu Wang, Xiaoqiang Zhou, Yingya Zhang, Luxin Yan, and Nong Sang.
\newblock Unianimate: Taming unified video diffusion models for consistent human image animation.
\newblock \emph{arXiv preprint arXiv:2406.01188}, 2024{\natexlab{b}}.

\bibitem[Wang et~al.(2024{\natexlab{c}})Wang, Zhang, Yuan, Qing, Gong, Zhang, Shen, Gao, and Sang]{wang2024recipe}
Xiang Wang, Shiwei Zhang, Hangjie Yuan, Zhiwu Qing, Biao Gong, Yingya Zhang, Yujun Shen, Changxin Gao, and Nong Sang.
\newblock A recipe for scaling up text-to-video generation with text-free videos.
\newblock In \emph{Proceedings of the IEEE/CVF Conference on Computer Vision and Pattern Recognition}, pages 6572--6582, 2024{\natexlab{c}}.

\bibitem[Wang et~al.(2023{\natexlab{d}})Wang, Chen, Ma, Zhou, Huang, Wang, Yang, He, Yu, Yang, et~al.]{wang2023lavie}
Yaohui Wang, Xinyuan Chen, Xin Ma, Shangchen Zhou, Ziqi Huang, Yi Wang, Ceyuan Yang, Yinan He, Jiashuo Yu, Peiqing Yang, et~al.
\newblock Lavie: High-quality video generation with cascaded latent diffusion models.
\newblock \emph{arXiv preprint arXiv:2309.15103}, 2023{\natexlab{d}}.

\bibitem[Wang et~al.(2024{\natexlab{d}})Wang, Li, Xie, Zhu, Guo, Dou, and Li]{wang2024customvideo}
Zhao Wang, Aoxue Li, Enze Xie, Lingting Zhu, Yong Guo, Qi Dou, and Zhenguo Li.
\newblock Customvideo: Customizing text-to-video generation with multiple subjects.
\newblock \emph{arXiv preprint arXiv:2401.09962}, 2024{\natexlab{d}}.

\bibitem[Wei et~al.(2023)Wei, Zhang, Ji, Bai, Zhang, and Zuo]{wei2023elite}
Yuxiang Wei, Yabo Zhang, Zhilong Ji, Jinfeng Bai, Lei Zhang, and Wangmeng Zuo.
\newblock Elite: Encoding visual concepts into textual embeddings for customized text-to-image generation.
\newblock In \emph{Proceedings of the IEEE/CVF International Conference on Computer Vision}, pages 15943--15953, 2023.

\bibitem[Wei et~al.(2024{\natexlab{a}})Wei, Zhang, Qing, Yuan, Liu, Liu, Zhang, Zhou, and Shan]{wei2024dreamvideo}
Yujie Wei, Shiwei Zhang, Zhiwu Qing, Hangjie Yuan, Zhiheng Liu, Yu Liu, Yingya Zhang, Jingren Zhou, and Hongming Shan.
\newblock Dreamvideo: Composing your dream videos with customized subject and motion.
\newblock In \emph{Proceedings of the IEEE/CVF Conference on Computer Vision and Pattern Recognition}, pages 6537--6549, 2024{\natexlab{a}}.

\bibitem[Wei et~al.(2024{\natexlab{b}})Wei, Zhang, Yuan, Wang, Qiu, Zhao, Feng, Liu, Huang, Ye, et~al.]{wei2024dreamvideo2}
Yujie Wei, Shiwei Zhang, Hangjie Yuan, Xiang Wang, Haonan Qiu, Rui Zhao, Yutong Feng, Feng Liu, Zhizhong Huang, Jiaxin Ye, et~al.
\newblock Dreamvideo-2: Zero-shot subject-driven video customization with precise motion control.
\newblock \emph{arXiv preprint arXiv:2410.13830}, 2024{\natexlab{b}}.

\bibitem[Wu et~al.(2024{\natexlab{a}})Wu, Li, Zeng, Zhang, Zhou, Li, Tong, and Chen]{wu2024motionbooth}
Jianzong Wu, Xiangtai Li, Yanhong Zeng, Jiangning Zhang, Qianyu Zhou, Yining Li, Yunhai Tong, and Kai Chen.
\newblock Motionbooth: Motion-aware customized text-to-video generation.
\newblock \emph{arXiv preprint arXiv:2406.17758}, 2024{\natexlab{a}}.

\bibitem[Wu et~al.(2024{\natexlab{b}})Wu, Zhang, Cun, Qi, Pu, Dou, Zheng, Shan, and Li]{wu2024videomaker}
Tao Wu, Yong Zhang, Xiaodong Cun, Zhongang Qi, Junfu Pu, Huanzhang Dou, Guangcong Zheng, Ying Shan, and Xi Li.
\newblock Videomaker: Zero-shot customized video generation with the inherent force of video diffusion models.
\newblock \emph{arXiv preprint arXiv:2412.19645}, 2024{\natexlab{b}}.

\bibitem[Wu et~al.(2024{\natexlab{c}})Wu, Zhang, Wang, Zhou, Zheng, Qi, Shan, and Li]{wu2024customcrafter}
Tao Wu, Yong Zhang, Xintao Wang, Xianpan Zhou, Guangcong Zheng, Zhongang Qi, Ying Shan, and Xi Li.
\newblock Customcrafter: Customized video generation with preserving motion and concept composition abilities.
\newblock \emph{arXiv preprint arXiv:2408.13239}, 2024{\natexlab{c}}.

\bibitem[Xu et~al.(2024{\natexlab{a}})Xu, Liu, Xing, Wang, Sun, Dan, Huang, Li, Cheng, Tai, et~al.]{xu2024facechain}
Chao Xu, Yang Liu, Jiazheng Xing, Weida Wang, Mingze Sun, Jun Dan, Tianxin Huang, Siyuan Li, Zhi-Qi Cheng, Ying Tai, et~al.
\newblock Facechain-imagineid: Freely crafting high-fidelity diverse talking faces from disentangled audio.
\newblock In \emph{Proceedings of the IEEE/CVF Conference on Computer Vision and Pattern Recognition}, pages 1292--1302, 2024{\natexlab{a}}.

\bibitem[Xu et~al.(2024{\natexlab{b}})Xu, Sun, Cheng, Wang, Liu, Sun, Huang, and Hauptmann]{xu2024combo}
Chao Xu, Mingze Sun, Zhi-Qi Cheng, Fei Wang, Yang Liu, Baigui Sun, Ruqi Huang, and Alexander Hauptmann.
\newblock Combo: Co-speech holistic 3d human motion generation and efficient customizable adaptation in harmony.
\newblock \emph{arXiv preprint arXiv:2408.09397}, 2024{\natexlab{b}}.

\bibitem[Xu et~al.(2017)Xu, Zhu, Choy, and Fei-Fei]{xu2017SGG-message-passing}
Danfei Xu, Yuke Zhu, Christopher~B Choy, and Li Fei-Fei.
\newblock Scene graph generation by iterative message passing.
\newblock In \emph{Proceedings of the IEEE conference on computer vision and pattern recognition}, pages 5410--5419, 2017.

\bibitem[Yang et~al.(2018)Yang, Lu, Lee, Batra, and Parikh]{yang2018graph-rcnn}
Jianwei Yang, Jiasen Lu, Stefan Lee, Dhruv Batra, and Devi Parikh.
\newblock Graph r-cnn for scene graph generation.
\newblock In \emph{Proceedings of the European conference on computer vision (ECCV)}, pages 670--685, 2018.

\bibitem[Yang et~al.(2025)Yang, Li, Zhao, Li, Xie, Tang, Lu, Liu, Zheng, Liu, and Yan]{CDT}
Nianzu Yang, Pandeng Li, Liming Zhao, Yang Li, Chen-Wei Xie, Yehui Tang, Xudong Lu, Zhihang Liu, Yun Zheng, Yu Liu, and Junchi Yan.
\newblock Rethinking video tokenization: A conditioned diffusion-based approach.
\newblock \emph{arXiv preprint arXiv:2503.03708}, 2025.

\bibitem[Yang et~al.(2024)Yang, Teng, Zheng, Ding, Huang, Xu, Yang, Hong, Zhang, Feng, et~al.]{yang2024cogvideox}
Zhuoyi Yang, Jiayan Teng, Wendi Zheng, Ming Ding, Shiyu Huang, Jiazheng Xu, Yuanming Yang, Wenyi Hong, Xiaohan Zhang, Guanyu Feng, et~al.
\newblock Cogvideox: Text-to-video diffusion models with an expert transformer.
\newblock \emph{arXiv preprint arXiv:2408.06072}, 2024.

\bibitem[Yatim et~al.(2024)Yatim, Fridman, Bar-Tal, Kasten, and Dekel]{yatim2024space}
Danah Yatim, Rafail Fridman, Omer Bar-Tal, Yoni Kasten, and Tali Dekel.
\newblock Space-time diffusion features for zero-shot text-driven motion transfer.
\newblock In \emph{Proceedings of the IEEE/CVF Conference on Computer Vision and Pattern Recognition}, pages 8466--8476, 2024.

\bibitem[Yuan et~al.(2022)Yuan, Jiang, Albanie, Feng, Huang, Ni, and Tang]{yuan2022rlip}
Hangjie Yuan, Jianwen Jiang, Samuel Albanie, Tao Feng, Ziyuan Huang, Dong Ni, and Mingqian Tang.
\newblock Rlip: Relational language-image pre-training for human-object interaction detection.
\newblock \emph{Advances in Neural Information Processing Systems}, 35:\penalty0 37416--37431, 2022.

\bibitem[Yuan et~al.(2024{\natexlab{a}})Yuan, Zhang, Wang, Wei, Feng, Pan, Zhang, Liu, Albanie, and Ni]{yuan2024instructvideo}
Hangjie Yuan, Shiwei Zhang, Xiang Wang, Yujie Wei, Tao Feng, Yining Pan, Yingya Zhang, Ziwei Liu, Samuel Albanie, and Dong Ni.
\newblock Instructvideo: Instructing video diffusion models with human feedback.
\newblock In \emph{Proceedings of the IEEE/CVF Conference on Computer Vision and Pattern Recognition}, pages 6463--6474, 2024{\natexlab{a}}.

\bibitem[Yuan et~al.(2024{\natexlab{b}})Yuan, Huang, He, Ge, Shi, Chen, Luo, and Yuan]{yuan2024identity}
Shenghai Yuan, Jinfa Huang, Xianyi He, Yunyuan Ge, Yujun Shi, Liuhan Chen, Jiebo Luo, and Li Yuan.
\newblock Identity-preserving text-to-video generation by frequency decomposition.
\newblock \emph{arXiv preprint arXiv:2411.17440}, 2024{\natexlab{b}}.

\bibitem[Zhang et~al.(2023{\natexlab{a}})Zhang, Wu, Liu, Zhao, Ran, Gu, Gao, and Shou]{show1}
David~Junhao Zhang, Jay~Zhangjie Wu, Jia-Wei Liu, Rui Zhao, Lingmin Ran, Yuchao Gu, Difei Gao, and Mike~Zheng Shou.
\newblock Show-1: Marrying pixel and latent diffusion models for text-to-video generation.
\newblock \emph{arXiv preprint arXiv:2309.15818}, 2023{\natexlab{a}}.

\bibitem[Zhang et~al.(2023{\natexlab{b}})Zhang, Rao, and Agrawala]{zhang2023adding}
Lvmin Zhang, Anyi Rao, and Maneesh Agrawala.
\newblock Adding conditional control to text-to-image diffusion models.
\newblock In \emph{Proceedings of the IEEE/CVF international conference on computer vision}, pages 3836--3847, 2023{\natexlab{b}}.

\bibitem[Zhang et~al.(2025)Zhang, Wang, Jiang, Fan, Xu, and Qi]{zhang2025fantasyid}
Yunpeng Zhang, Qiang Wang, Fan Jiang, Yaqi Fan, Mu Xu, and Yonggang Qi.
\newblock Fantasyid: Face knowledge enhanced id-preserving video generation.
\newblock \emph{arXiv preprint arXiv:2502.13995}, 2025.

\bibitem[Zhao et~al.(2023)Zhao, Gu, Wu, Zhang, Liu, Wu, Keppo, and Shou]{motionDirector}
Rui Zhao, Yuchao Gu, Jay~Zhangjie Wu, David~Junhao Zhang, Jiawei Liu, Weijia Wu, Jussi Keppo, and Mike~Zheng Shou.
\newblock Motiondirector: Motion customization of text-to-video diffusion models.
\newblock \emph{arXiv preprint arXiv:2310.08465}, 2023.

\bibitem[Zheng et~al.(2024)Zheng, Peng, Yang, Shen, Li, Liu, Zhou, Li, and You]{Open-Sora}
Zangwei Zheng, Xiangyu Peng, Tianji Yang, Chenhui Shen, Shenggui Li, Hongxin Liu, Yukun Zhou, Tianyi Li, and Yang You.
\newblock Open-sora: Democratizing efficient video production for all, 2024.
\newblock https://github.com/hpcaitech/Open-Sora.

\bibitem[Zhou et~al.(2024{\natexlab{a}})Zhou, Zhang, Gu, Zhao, Shi, and Sun]{zhou2024sugar}
Yufan Zhou, Ruiyi Zhang, Jiuxiang Gu, Nanxuan Zhao, Jing Shi, and Tong Sun.
\newblock Sugar: Subject-driven video customization in a zero-shot manner.
\newblock \emph{arXiv preprint arXiv:2412.10533}, 2024{\natexlab{a}}.

\bibitem[Zhou et~al.(2024{\natexlab{b}})Zhou, Zhou, Cheng, Feng, and Hou]{zhou2024storydiffusion}
Yupeng Zhou, Daquan Zhou, Ming-Ming Cheng, Jiashi Feng, and Qibin Hou.
\newblock Storydiffusion: Consistent self-attention for long-range image and video generation.
\newblock \emph{arXiv preprint arXiv:2405.01434}, 2024{\natexlab{b}}.

\end{thebibliography}
}

\clearpage

\maketitlesupplementary
\appendix
\section{Appendix}
\subsection{Experimental Setup}
\label{app:exp_setup}
\noindent\textbf{Datasets.}\quad
We select 26 types of human interaction videos from the NTU RGB+D Action Recognition Dataset~\cite{shahroudy2016ntu, liu2019ntu} for training. The names of these interactions and their annotated textual descriptions are provided in Tab.~\ref{tab:relation_name_list}.

\noindent\textbf{Baselines.}\quad
Due to the current lack of relational video customization methods, we consider four baselines and detail the implementation of each method below:
\textit{1)} Base Model Mochi~\cite{genmo2024mochi}. We input the test text prompts into the original Mochi for inference and evaluate the results.
\textit{2)} Direct LoRA Fine-tuning. We insert LoRAs into all the Query, Key, Value matrices, and FFNs in Mochi for training and inference. The training iterations are set to 1,000. Other training settings, such as the optimizer and LoRA rank, are the same as those in our \frameworkplain.
\textit{3)} ReVersion~\cite{huang2024reversion}. As ReVersion is designed for relational image customization and cannot be directly applied for video generation, we adapt ReVersion to the base model Mochi based on their official code\footnote{\url{https://github.com/ziqihuangg/ReVersion}}. The training settings follow the default settings provided in the official ReVersion paper.
\textit{4)} MotionInversion~\cite{wang2024motioninversion}. 
Given that MotionInversion is designed based on the Temporal Attention layers within the UNet architecture, and such layers are absent in the MM-DiT architecture, we adapt MotionInversion to Mochi using their official code\footnote{\url{https://github.com/EnVision-Research/MotionInversion}}.
Specifically, we integrate the two embeddings from MotionInversion into the query, key, and value matrices of full attention, adhering to their official paper. The learning rate is set to 2e-4, and the weight decay is set to 0.01. The training iterations are 3,000, with other settings consistent with our method. During inference, we utilize the differencing operation from their official paper to mitigate the appearance biases in motion embeddings.

\noindent\textbf{Evaluation metrics.}\quad
We detail the proposed Relation Accuracy metric utilizing Vision-Language Models (VLMs). Specifically, we input all generated videos into Qwen-VL-Max~\cite{Qwen-VL}, the state-of-the-art Visual Question Answering (VQA) model, to determine if the generated video conforms to the specified relation, prompting it to return either ``yes'' or ``no.''
Directly inputting an entire 61-frame video into the VLM would require significant resources and slow response times. To address this, we evenly extract five key frames from each video, including the first and last two frames, and input them into the VLM.
The text input template for the VLM is: ``Based on the keyframes of the video, analyze whether the two subjects are performing human-like \{\} interactions. The answer should be 'yes' or 'no'." The ``\{\}'' is replaced with a specific relation name, such as ``handshaking'', for evaluation.
We test all videos ten times, count the responses for all videos, convert these into percentages of relation accuracy, and compute the average accuracy as the Relation Accuracy score.

\subsection{More Results}
\label{app:more_results}
\noindent\textbf{Details about the user study.}\quad
\label{app:user_study}
We conduct a user study involving 180 groups of videos with 15 randomly selected relations. Participants are presented with three sets of questions for each of the four anonymous methods, paired with a reference video and a textual prompt.
For each group of four generated videos, participants are asked the following questions:
(1) Relation Alignment:
``Which interaction exhibited in videos is more consistent with the reference video?'';
(2) Text Alignment:
``Which video better matches the text description?'';
and (3) Overall Quality:
``Which video exhibits better quality and minimal flicker?''.
The results of the user study are illustrated in Fig.~\ref{fig:exp_attn_map_and_user_study}(b).

\begin{figure*}[t]
  \centering
   \includegraphics[width=1.0\linewidth]{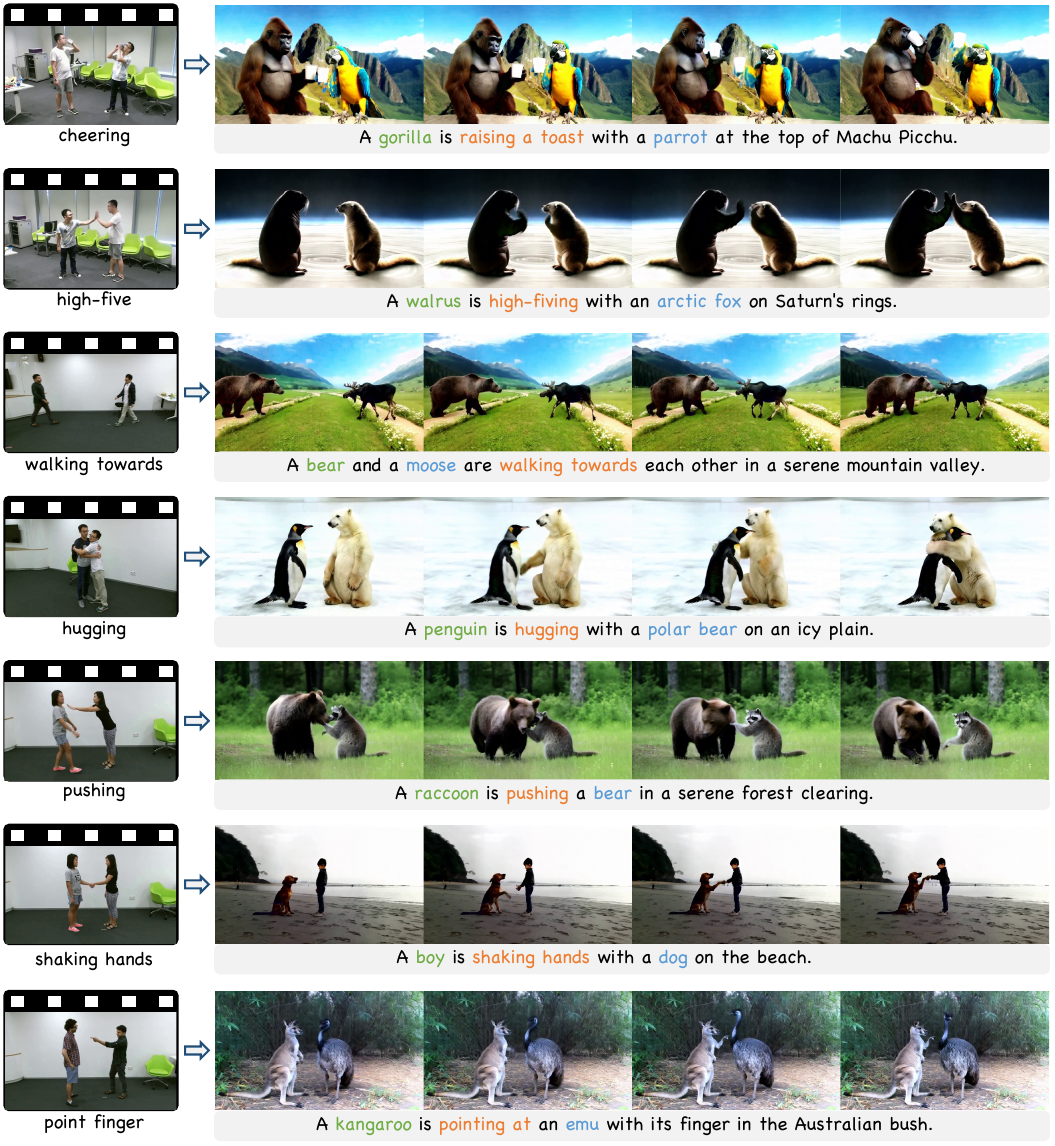}
   \caption{\textbf{More qualitative results of \frameworkplain (1/2)}. Please zoom in for a better view.}
   \label{fig:more_result_1}
\end{figure*}

\begin{figure*}[t]
  \centering
   \includegraphics[width=1.0\linewidth]{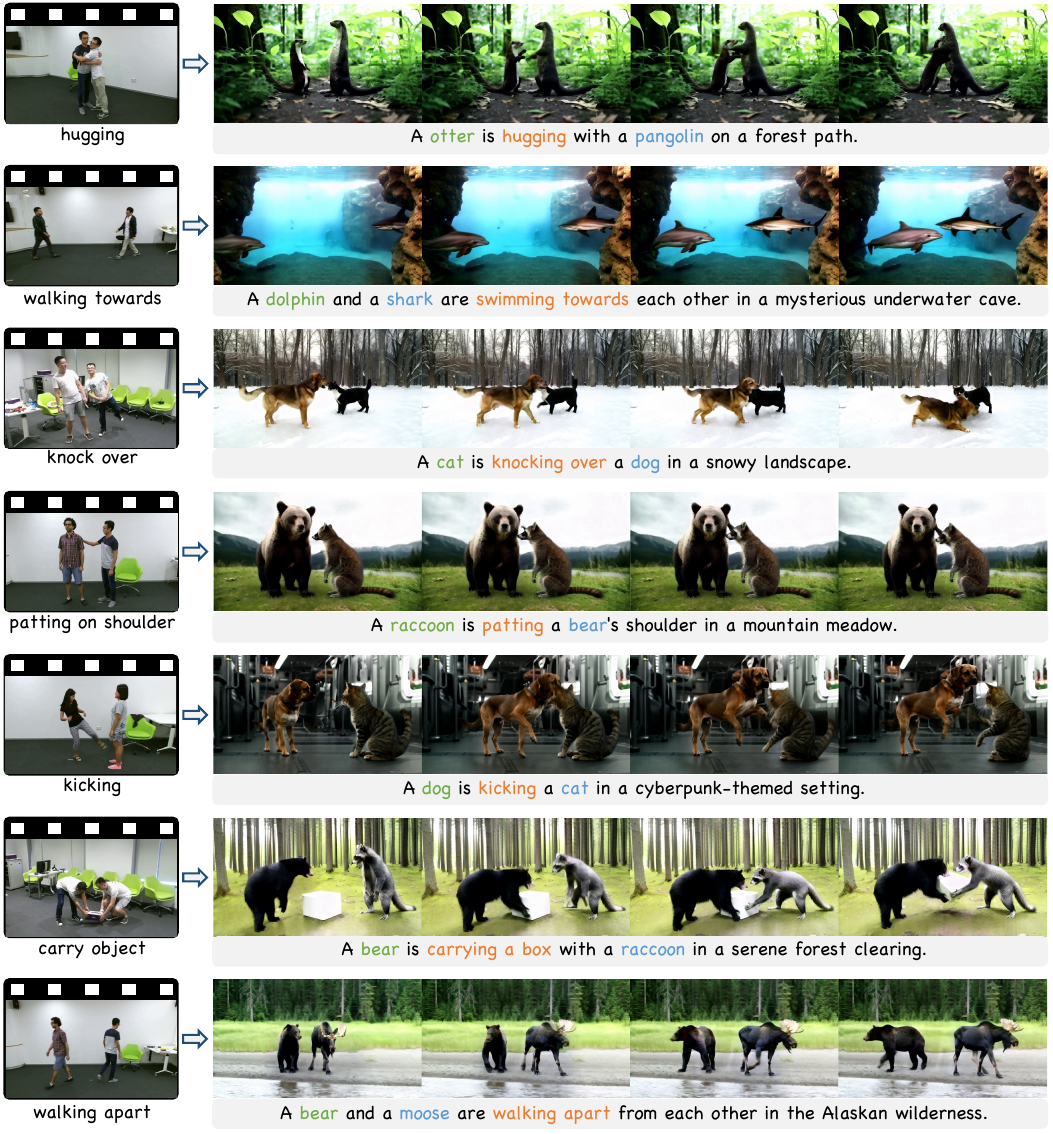}
   \caption{\textbf{More qualitative results of \frameworkplain (2/2)}. Please zoom in for a better view.}
   \label{fig:more_result_2}
\end{figure*}

\noindent\textbf{More qualitative results.}\quad
To further demonstrate the effectiveness of our \frameworkplain, we present additional visual results in Figs.~\ref{fig:more_result_1} and~\ref{fig:more_result_2}. These examples illustrate the capability of our method to generate videos that align with the specified relations and textual descriptions.

\subsection{More Ablation Studies}
\label{app:more_ablation}
\noindent\textbf{Effects of Loss Lam $\lambda_{1}$.}\quad
To determine the optimal value for the loss weight $\lambda_{1}$, we vary its value and measure its impact.
As shown in Tab.~\ref{tab:ablation_loss_lam}, increasing the loss weight of \CLName results in degradation of Relation Accuracy.
We argue that over-emphasizing contrastive learning may ignore detailed information from training videos, leading to degraded performance.
Therefore, we set $\lambda_{1}$ to 0.01 for the best performance.
\begin{table}[h]
    \centering
    \footnotesize
    \begin{tabular}{ccccc} 
        $\lambda_{1}$ & \tabincell{c}{\textbf{Relation}\\\textbf{Accuracy}} & 
        \textbf{CLIP-T} & \tabincell{c}{\textbf{Temporal}\\\textbf{Consistency}} & 
        \textbf{FVD$\downarrow$} \\ 
        \shline
        \noalign{\smallskip}
         0.01 & \textbf{0.4452}\std{$\pm$0.01} & \underline{0.3248} & \textbf{0.9954} & \underline{2079.87} \\
         0.10 & \underline{0.3964}\std{$\pm$0.03} & 0.3241 & \textbf{0.9954} & 2088.71 \\
         1.00 & 0.2998\std{$\pm$0.01} & \textbf{0.3254} & \textbf{0.9954} & \textbf{1971.29} \\
    \end{tabular}
    \caption{\textbf{Ablation study of the loss weight $\lambda_1$.}
    }
    \label{tab:ablation_loss_lam}
\end{table}

\noindent\textbf{Effects of Mask Lam.}\quad
To identify the optimal mask weight $\lambda_{m}$, we explore various values and assess their impact.
As shown in Tab.~\ref{tab:ablation_mask_lam}, both excessively high and low mask weights can result in poor performance. We argue that low mask weights fail to direct the model's focus on the area of interest, while high weights lead to excessive emphasis, causing the neglect of other visual cues. Based on the results, we set $\lambda_{m}$ to 50.
\begin{table}[h]
    \centering
    \footnotesize
    \begin{tabular}{ccccc} 
        $\lambda_{m}$ & \tabincell{c}{\textbf{Relation}\\\textbf{Accuracy}} & 
        \textbf{CLIP-T} & \tabincell{c}{\textbf{Temporal}\\\textbf{Consistency}} & 
        \textbf{FVD$\downarrow$} \\ 
        \shline
        \noalign{\smallskip}
         1  & 0.3469\std{$\pm$0.07} & 0.2826 & 0.9942 & 2294.98 \\
         25 & 0.3899\std{$\pm$0.04} & 0.3185 & \underline{0.9953} & 2117.49 \\
         50 & \textbf{0.4452}\std{$\pm$0.01} & \textbf{0.3248} & \textbf{0.9954} & \underline{2079.87} \\
         100 & \underline{0.4018}\std{$\pm$0.04} & \underline{0.3246} & 0.9952 & \textbf{2050.10} \\
    \end{tabular}
    \caption{\textbf{Ablation study of the mask weight $\lambda_{m}$.}
    }
    \label{tab:ablation_mask_lam}
\end{table}

\noindent\textbf{Effects of positive and negative numbers.}\quad
We conduct ablation studies to investigate the effects of varying the number of positive and negative samples in \CLName. A higher number of positive samples emphasizes the alignment of relational information during training, while an increased number of negative samples focuses more on distinguishing appearance information. We observe that different combinations have varying effects, and based on the experimental results, we chose to set $n_\text{pos}$ to 4 and $n_\text{neg}$ to 10.
\begin{table}[h]
    \centering
    \footnotesize
    \begin{tabular}{cc|cccc} 
        $n_\textbf{pos}$ & $n_\textbf{neg}$ & \tabincell{c}{\textbf{Relation}\\\textbf{Accuracy}} &
        \textbf{CLIP-T} & \tabincell{c}{\textbf{Temporal}\\\textbf{Consistency}} & 
        \textbf{FVD$\downarrow$} \\ 
        \shline
        \noalign{\smallskip}
        1 & 10 & 0.3151\std{$\pm$0.04} & \textbf{0.3259} & \underline{0.9954} & 2089.79 \\
        1 & 30 & 0.2817\std{$\pm$0.03} & 0.3125 & \textbf{0.9957} & \underline{2067.35} \\
        1  & 60 & 0.3338\std{$\pm$0.06} & 0.3154 & \underline{0.9954} & 2113.27 \\
        2 & 10 & 0.3321\std{$\pm$0.02} & 0.3227 & 0.9950 & 2254.62 \\
        4 & 10 & \textbf{0.4452}\std{$\pm$0.01} & \underline{0.3248} & \underline{0.9954} & 2079.87 \\
        2 & 30 & \underline{0.4378}\std{$\pm$0.02} & 0.3168 & 0.9953 & \textbf{2009.92} \\
        4 & 60 & 0.3793\std{$\pm$0.03} & 0.3237 & 0.9952 & 2156.28 \\
    \end{tabular}
    \caption{\textbf{Ablation study of the number of positive and negative samples.}
    }
    \label{tab:ablation_pos_neg_num}
\end{table}

\begin{table*}[ht]
    \caption{The list of 26 human interactions with their textual prompts.}
    \centering
    \small
    \begin{tabular}{rl}
    \shline
    1.& \textbf{walking apart}: ``A person and a person are walking apart from each other.''\\
    2.& \textbf{walking towards}: ``A person and a person are walking towards each other.''\\
    3.& \textbf{shaking hands}: ``A person is shaking hands with a person.''\\
    4.& \textbf{hugging}: ``A person is hugging with a person.''\\
    5.& \textbf{point finger}: ``A person is pointing at a person with his finger.''\\
    6.& \textbf{pat on back}: ``A person is patting a person's shoulder.''\\
    7.& \textbf{pushing}: ``A person is pushing a person.''\\
    8.& \textbf{kicking}: ``A person is kicking a person.''\\
    9.& \textbf{punch or slap}: ``A person is punching a person.''\\
    10.& \textbf{rock-paper-scissors}: ``A person is playing rock-paper-scissors with a person.''\\
    11.& \textbf{support somebody}: ``A person is supporting a person while walking.''\\
    12.& \textbf{whisper}: ``A person is whispering to a person.''\\
    13.& \textbf{follow}: ``A person is following a person.''\\
    14.& \textbf{take a photo}: ``A person is taking a photo of a person.''\\
    15.& \textbf{carry object}: ``A person is carrying a box with a person.''\\
    16.& \textbf{cheers and drink}: ``A person is raising a toast with a person.''\\
    17.& \textbf{high-five}: ``A person is high-fiving with a person.''\\
    18.& \textbf{step on foot}: ``A person is stepping on a person's foot.''\\
    19.& \textbf{shoot with gun}: ``A person is shooting a person with a water gun.''\\
    20.& \textbf{knock over}: ``A person is knocking over a person.''\\
    21.& \textbf{giving object}: ``A person is giving an object to a person.''\\
    22.& \textbf{touch pocket}: ``A person is touching a person's pocket.''\\
    23.& \textbf{hit with object}: ``A person is hitting a person with an object.''\\
    24.& \textbf{wield knife}: ``A person is wielding a toy knife towards a person.''\\
    25.& \textbf{grab stuff}: ``A person is grabbing an item from a person.''\\
    26.& \textbf{exchange things}: ``A person and a person are exchanging items with each other.''\\
    \shline
    \end{tabular}
    \label{tab:relation_name_list}
\end{table*}

\end{document}